\documentclass[runningheads]{llncs}
\usepackage{graphicx}
\usepackage{tikz}
\usepackage{comment}
\usepackage{amsmath,amssymb} % define this before the line numbering.
\usepackage{color}
\usepackage{booktabs}
\usepackage{mathtools,amssymb}
\usepackage{footnote}
\usepackage{diagbox}
\usepackage{caption}
\usepackage{subcaption}
\usepackage{multirow}
\usepackage{tabularx}
\usepackage{amsfonts}
\usepackage{siunitx}
\usepackage{amsmath}
\usepackage{algorithm}
\usepackage{algpseudocode}
\usepackage{tabularx,booktabs}

\usepackage[accsupp]{axessibility}  % Improves PDF readability for those with disabilities.

\def\B#1{\textbf{#1}} 
\newcolumntype{L}[1]{>{\raggedright\let\newline\\\arraybackslash\hspace{0pt}}m{#1}}
\newcolumntype{C}[1]{>{\centering\let\newline\\\arraybackslash\hspace{0pt}}m{#1}}
\newcolumntype{R}[1]{>{\raggedleft\let\newline\\\arraybackslash\hspace{0pt}}m{#1}}

\usepackage[capitalize]{cleveref}
\crefname{section}{Sec.}{Secs.}
\Crefname{section}{Section}{Sections}
\Crefname{table}{Table}{Tables}
\crefname{table}{Tab.}{Tabs.}

\begin{document}
% \linenumbers
\pagestyle{headings}
\mainmatter
\def\ECCVSubNumber{}  % Insert your submission number here

\title{Knowledge-Spreader: Learning Facial Action Unit Dynamics with Extremely Limited Labels} % Replace with your title

%\begin{comment}
\titlerunning{ \ECCVSubNumber} 
\authorrunning{ \ECCVSubNumber} 
\author{Xiaotian Li, Xiang Zhang, Taoyue Wang, Lijun Yin}
\institute{State University of New York at Binghamton \ECCVSubNumber}
%\end{comment}
%******************

\begin{comment}
\titlerunning{Abbreviated paper title}
\author{First Author\inst{1}\orcidID{0000-1111-2222-3333} \and
Second Author\inst{2,3}\orcidID{1111-2222-3333-4444} \and
Third Author\inst{3}\orcidID{2222--3333-4444-5555}}
%
\authorrunning{F. Author et al.}
\institute{Princeton University, Princeton NJ 08544, USA \and
Springer Heidelberg, Tiergartenstr. 17, 69121 Heidelberg, Germany
\email{lncs@springer.com}\\
\url{http://www.springer.com/gp/computer-science/lncs} \and
ABC Institute, Rupert-Karls-University Heidelberg, Heidelberg, Germany\\
\email{\{abc,lncs\}@uni-heidelberg.de}}
\end{comment}
%******************
\maketitle

\begin{abstract}
Recent studies on the automatic detection of facial action unit (AU) have extensively relied on large-sized annotations. However, manually AU labeling is difficult, time-consuming, and costly. Most existing semi-supervised works ignore the informative cues from the temporal domain, and are highly dependent on densely annotated videos, making the learning process less efficient. To alleviate these problems, we propose a deep semi-supervised framework \textbf{Knowledge-Spreader (KS)}, which differs from conventional methods in two aspects. First, rather than only encoding human knowledge as constraints, KS also learns the Spatial-Temporal AU correlation knowledge in order to strengthen its out-of-distribution generalization ability. Second, we approach KS by applying consistency regularization and pseudo-labeling in multiple student networks alternately and dynamically. It spreads the spatial knowledge from labeled frames to unlabeled data, and completes the temporal information of partially labeled video clips. Thus, the design allows KS to learn AU dynamics from video clips with only \textbf{one label} allocated, which significantly reduce the requirements of using annotations. Extensive experiments demonstrate that the proposed KS achieves competitive performance as compared to the state of the arts under the circumstances of using only \textbf{2\%} labels on BP4D and \textbf{5\%} labels on DISFA. In addition, we test it on our newly developed large-scale comprehensive emotion database, which contains considerable samples across well-synchronized and aligned sensor modalities for easing the scarcity issue of annotations and identities in human affective computing. The code and new database will be released to the research community.
\end{abstract}

% We approach KS by applying consistency regularization and pseudo-labeling in multiple student networks alternately and dynamically, spreading the spatial knowledge from labeled frames to unlabeled frames, and completing the temporal information of single-frame-labeled video clips.

% KS spreads the spatial knowledge from discretely-labeled data to unlabeled frames by 

% The key of KS is the dynamic spread of the spatial knowledge from labeled frames to unlabeled frames using both consistency , 

% completing the temporal information of partially-labeled video clips, 

% and then distilling the temporal information back to the discretely-labeled data in order to achieve an optimal fusion of Spatial-Temporal information. 

\section{Introduction}

\label{sec:intro}

% Overview of the proposed Knowledge-Spreader. To reduce the annotation redundancy, only one label is allocated for every video clip ($k$ frames). Two branches of networks (F and V) are set to learn the two level AU correlation information (spatial and temporal) respectively. Different from conventional knowledge distillation, we dynamically switch the student network of B from $S^{1}_{b}$ to $S^{n}_{b}$ with the labeled frame shifting for input clips. After spreading the spatial knowledge (red arrow) of AUs semantic correlations to every semi-supervised student networks $\left \{ S^{1}_{b}, S^{2}_{b}, ..., S^{n}_{b} \right \}$ in V. Afterwards, the spatial features extracted by $\left \{ S^{1}_{b}, S^{2}_{b}, ..., S^{n}_{b} \right \}$ can be integrated for learning the temporal knowledge by student network $T_{b}$ and spread back (blue arrow) to F for robust fusing of spatial and temporal knowledge. Noting that the pseudo-labeling (blue square) is applied for learning with unlabeled data.

Understanding human facial action dynamics is crucial for human-computer interaction (HCI), which reflects the individuals’ affective states to enhance the affinity of communication. Facial action unit (AU) detection plays a vital role in automatic facial action analysis. Over the past few years, a substantial amount of works~\cite{Introduction_5} ~\cite{Introduction_6} ~\cite{Introduction_7} based on deep learning have shown the superior power of constructing informative features against the traditional methods. The majority of the recent advances~\cite{Relatedworks_13} ~\cite{Introduction_4} ~\cite{Experiment_5} ~\cite{Relatedworks_6} ~\cite{Introduction_1} ~\cite{Introduction_2} ~\cite{Introduction_3} in this area are heavily relied on large-scale labeled datasets for achieving remarkable learning performance. However, a lab-controlled AU video typically contains thousands of frames that need to be densely labeled by human annotators. As a result, datasets for dynamic facial action detection suffer high redundancy both content-wise and annotation-wise.

There are only limited works attempting to mitigate the demand for dense AU occurrence annotations. For example, the researchers~\cite{Relatedworks_11} ~\cite{Relatedworks_12} ~\cite{Introduction_8} ~\cite{Introduction_10} summarize the distribution from existing ground-truth AU labels as the prior for detecting AUs with partially labeled data. However, the approaches of applying the distribution prior may fall into sub-optimal due to lacking adaptation mechanism. Deep semi-supervised learning has become a new choice to overcome these issues for their powerful representation and generalization ability. The existing deep semi-supervised learning methods can be roughly categorised into three methods (i.e. generative method, consistency regularization method, and pseudo-labeling methods). Recently, some hybrid methods attract the attention from many researchers. MixMatch~\cite{Introduction_14} combines entropy minimization and consistency regularization in an unified loss function. FixMatch~\cite{Experiment_4} combines consistency regularization and pseudo-labeling considering both labeled and unlabeled data should be trained simultaneously. Whereas, applying these image-level models to dynamic datasets is challenging due to some nuisance factors such as motion blur, video defuse, and frequent pose occlusions. Besides, human facial action characteristics present strong dependencies and mutual exclusive relation among spatially local regions~\cite{Introduction_12}. For instance, the AU1 and AU2 are usually co-existence due to the constraint of facial muscles. Some recent works~\cite{Introduction_13} ~\cite{Relatedworks_8} prove that the relationship of different AUs in temporal domain is also an important factor for robust AU detection. Both of the two factors have been studied for fully supervised methods, but there lacks a framework to explore if learning both spatial and temporal AU correlation knowledge can boost the semi-supervised process.

% A recent work~\cite{Relatedworks_17} has been proved that using a few sparsely sampled video clips for action recognition is sometimes more accurate and efficient than using densely extracted offline features from full-length videos. It inspires us to explore whether this principle is practical in video AU detection. 
\begin{figure}
\begin{center}
\includegraphics[width=0.90\linewidth]{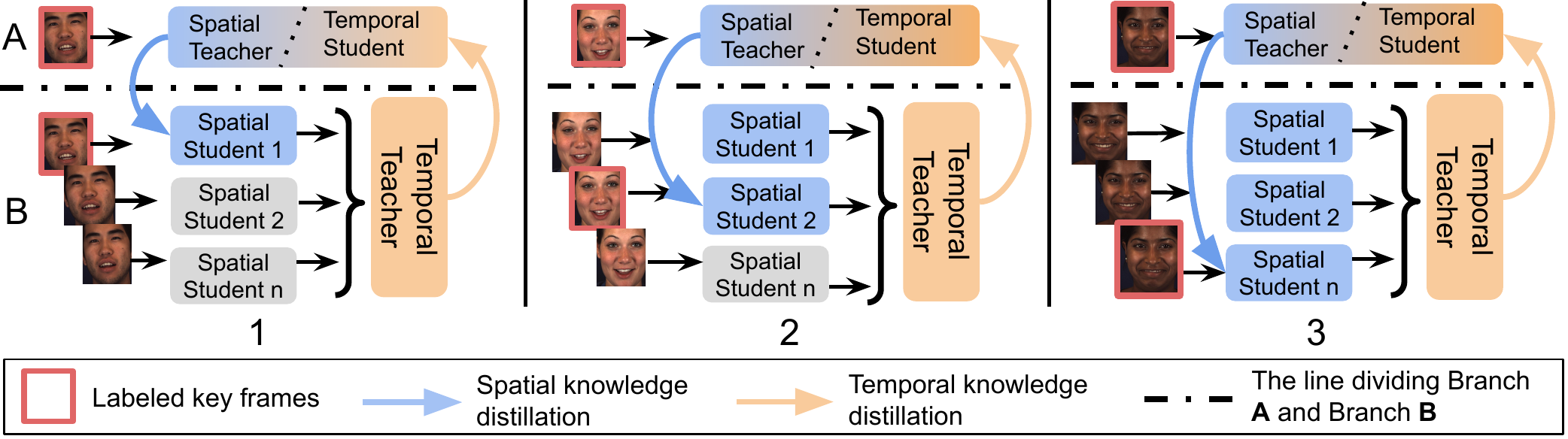}
    \caption{Overall pipeline of Knowledge Spreading. (1) Train a Spatial Teacher for learning supervised image-level spatial knowledge in Branch $A$ (\emph{qualitative training}), and train a group of Spatial Students for semi-supervised learning in Branch $B$ (\emph{quantitative training}). (2) Process spatial knowledge knowledge distillation~\cite{Methodology_8} when one of the Spatial Students gets labeled key frames (red square) as inputs while other Spatial Students infer pseudo labels using unlabeled frames. With the position of key frame shifting from 1 to 3 repeatedly, the spatial AU knowledge (blue) will be gradually spread to every Spatial Student and boost the \emph{quantitative training} of pseudo-labeling. (3) By completing and encoding the outputs of all Spatial Students, Temporal Teacher is deployed to learn the AU temporal cues (orange). (4) Distill it back to Temporal Student in A for getting an optimal fusion of Spatial-Temporal information.}
\label{fig:pipeline_module}
\end{center}
\end{figure}

In this paper, we propose an end-to-end trainable framework \textbf{Knowledge-Spreader} (KS) for learning semi-supervised facial action dynamics. Knowledge-Spreader improves conventional semi-supervised learning in three ways. First, different from previous video-level works that need fully supervisory information or at least a few labeled frames as input segments, KS just needs limited sparsely labeled video clips, with only one label allocated. This alleviates the demand on densely labeled data effectively. Second, incorporating the weak supervisory information from both human prior and AU correlation knowledge makes KS more efficient and robust for inferring out-of-distribution information. KS learns both spatial and temporal AU correlation knowledge by transformers. The powerful generalization capability of deep learning can reduce some side effects caused by incorrect knowledge summarized by humans. For instance, some AU intensity estimation works~\cite{Introduction_16} assume that there are only one peak or valley in a video sequence, while ignore the special situation that multiple peaks and valleys exist in a video. Third, our work improves the hybrid semi-supervised method (consistency regularization and pseudo-labeling) to a video level model by introducing an operation ``Knowledge Spreading''. The overall pipeline of Knowledge Spreading is shown in \Cref{fig:pipeline_module}. The \emph{qualitative training} in \Cref{fig:pipeline_module} refers that the network is trained with a small amount of labeled images, while the \emph{quantitative training} indicates the training with a large-sized unlabeled data. Our contribution lies in three-fold: (1) We propose a deep semi-supervised architecture for AU detection by jointly utilizing the learned Spatial-Temporal AU correlation knowledge and human knowledge. (2) This is the first work to incorporate knowledge spreading by dynamic knowledge distillation and temporal confirmed pseudo-label in a self-supervised manner, to learn spatio-temporal information using extremely limited single-frame labeled video clips. (3) We have built a new spontaneous emotion database by capturing 3D geometric facial sequences, 2D facial videos, thermal videos, and physiological data sequences from 233 participants across two years period. The new database will be released to the research community along with the paper being published.

\section{Related Work}

\subsection{Automatic Facial AU Detection}
% We provide a brief review of related works in deep-learning based facial AU detection. 

% Most of existing works address the issues from three aspects: (1) Learning region of interest (ROI) (2) Learning AU correlation (3) Alleviating the demand for AU annotation. 

Different from general computer vision tasks, facial AUs are defined to be associated with atomic local facial muscles, which in turn correspond to the appearance features of different regions in the face. Thus, previous research~\cite{Relatedworks_4} ~\cite{Relatedworks_2} ~\cite{Experiments_8} ~\cite{Introduction_4} ~\cite{Relatedworks_5} has extensively studied how to use manually-defined regions, local patches, facial landmarks, heatmap, attention mechanism, and other methods to localize detailed facial parts. However, considering the structural information and dependencies among different AUs, conventional CNN is criticized to be incapable of fully characterizing their correlations. Recent advances~\cite{Relatedworks_6} ~\cite{Relatedworks_7} ~\cite{Introduction_2} ~\cite{Introduction_3} ~\cite{Relatedworks_20} start to explore the AU relations by using data distribution, conditional random field, Bayesian Network, graph neural networks, attention mechanism, and other approaches. ~\cite{Relatedworks_8} ~\cite{Relatedworks_9} ~\cite{Relatedworks_10} attempt to integrate the inter-relationship of AUs on both spatial and temporal dimensions for learning more robust AU dependencies.

Some of the existing semi-supervised or self-supervised methods~\cite{Relatedworks_13} ~\cite{Introduction_4} ~\cite{Experiment_5} ~\cite{Relatedworks_6} still rely on large-sized annotations and extra unlabeled data. Only a small amount of research has attempted to ease the intensive demand for AU annotations. ~\cite{Relatedworks_11}  and~\cite{Relatedworks_12} summarize the distribution from ground-truth AU labels as the prior for detecting AUs with limited labeled data with semi-supervised learning.~\cite{Introduction_8} ~\cite{Introduction_10} leverage the joint distribution of features, AUs, and facial expression to explore the global dependencies among them. Different from all aforementioned works, KS can learn the informative Spatial-Temporal cues from sparse annotations by incorporating dynamic knowledge distillation and pseudo label.

\subsection{Semi-supervised learning}
 Semi-supervised learning is an approach to combine a few labeled data with a large amount of unlabeled data during training. Deep semi-supervised learning is a fast-growing filed with a wide broad of practical applications. FixMatch~\cite{Experiment_4} has been proved to be a simple but efficient SSL method by combining both consistency regularization and pseudo-labeling. Noisy Student improves the idea of self-training and distillation with the use of noise added to the student networks. However, these works are incapable to capture the facial action dynamics as an image-level model. Recent works~\cite{Experiment_1} ~\cite{Methodology_5} ~\cite{Relatedworks_21} propose to recognize actions from only a handful of sparsely labeled videos with the rich supervisory information in a semi-supervised way. Unlike these approaches that still need continuous and dense annotation, we aim to allocate only single-frame label for each video clip to handle the condition that the labels are extremely sparse, missing, or scarce.

% Although the ``less-is-more'' principle is well addressed by~\cite{Relatedworks_17}, it is still challenging to apply it directly to multi-label AU recognition task considering the its precise labeling at the frame level. In another word, the discarded clips may contain some informative features we need to reserve. In this paper, we sparsely samples one annotation for every $k$ frames from full-length videos instead. We evaluate how the difference affects the performance of the proposed method in Sec.

% Bertasius \textit{et al}.~\cite{Relatedworks_18} provide an alternative solution that they leverage training videos with sparse annotations (every $N$ frames) to perform dense temporal pose propagation and estimation. Specifically, they use deformable convolutions to implicitly learn the pose warping between a labeled frame A and unlabeled frame B. The potential issue is the lacking of continuity between labeled frame and unlabeled frame which may lead it into sub-optimal. In this paper, we adopt the trading-off version of~\cite{Relatedworks_17} and~\cite{Relatedworks_18}. We leverage the sparse annotations (every $k$ frames) in training videos for reducing the redundancy issue of densely labeled AU videos. At the same time, the unlabeled neighbour frames are fully utilized to provide sufficient temporal cues by incorporating knowledge distillation, semi-supervised learning.

%------------------------------------------------------------------------
\section{Methodology}
\subsection{Overview}
\begin{figure*}[ht]
\begin{center}
\includegraphics[width=0.8\linewidth]{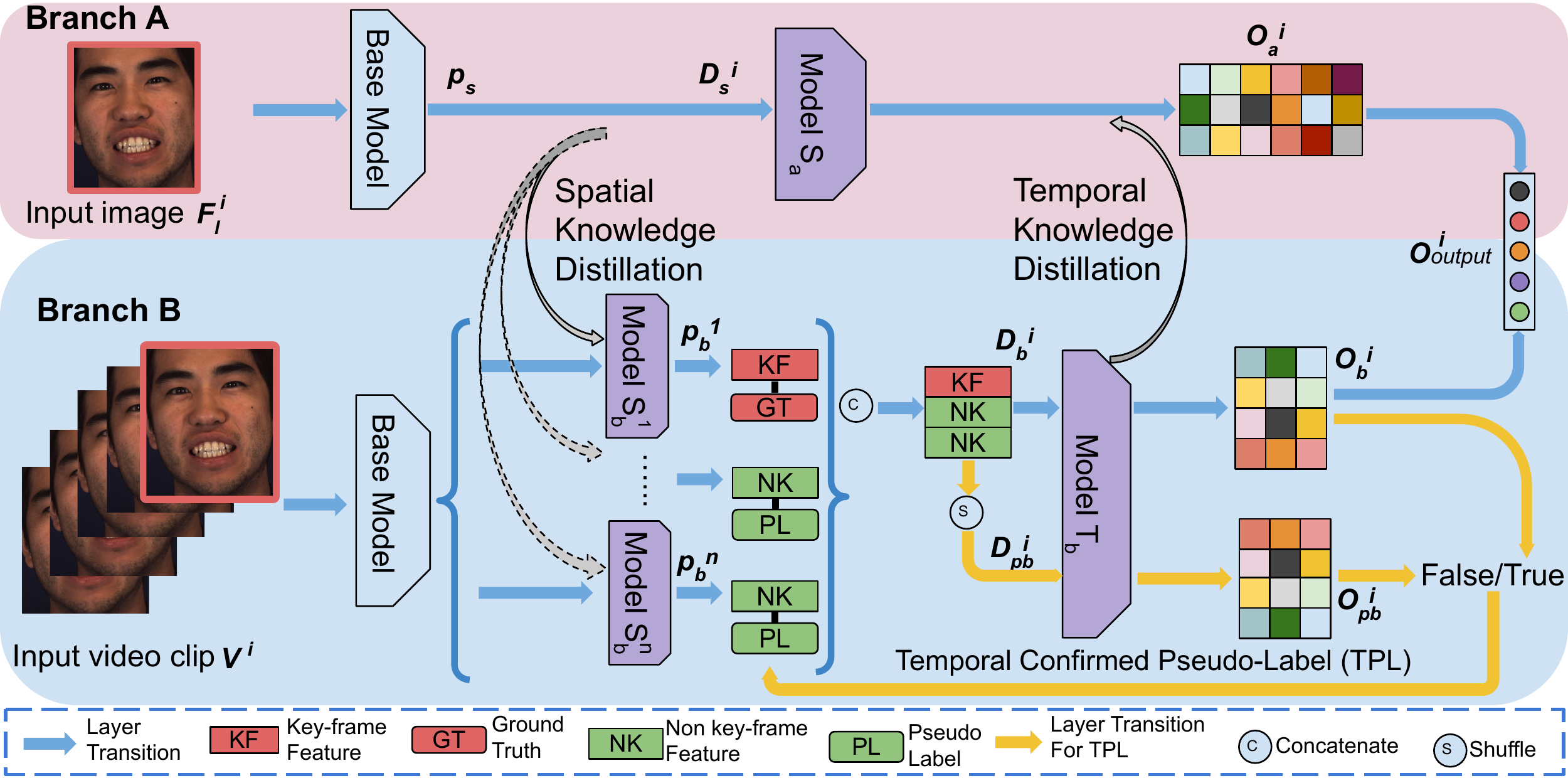}
\caption{\textbf{Illustration of the detailed architecture for Knowledge-Spreader.} The framework is consisted of two branches without sharing weights for learning image-level and video-level features respectively. Model $S_{a}$, $S_{b}$, and $T_{b}$ (purple) are networks used for learning Spatial-Temporal AU correlation knowledge. Two level distillation (grey) and pseudo-labeling (green) consist the proposed ``knowledge spreading''. A self-supervised module ``temporal confirmed pseudo-label'' (yellow) is designed for sensing the temporal perturbation and selecting pseudo labels with high confidence. Details are described in \Cref{sect:pd-confirmation}.
}
\label{fig:Overview_module}
\end{center}
\end{figure*}

Our goal is to detect facial action units with video clips with single-frame labels. Specifically, we assume the facial action labels in training data are available every $k$ frames. The $i$th video clip consisted of $n$ frames is a set $V^{i}=\left \{ F_{u,1}^{i},F_{u,2}^{i},\cdot \cdot \cdot ,F_{l,m}^{i},\cdot \cdot \cdot ,F_{u,n}^{i} \right \}$ where $F_{u}^{i}$ means unlabeled frame and $F_{l}^{i}$ means labeled frame. Knowledge-Spreader is consisted of three key modules: (1) Spatial-Temporal information learning module (SIL), (2) knowledge spreading module (KSM), and (3) temporal confirmed pseudo-label module (TPL). The network structure of KS is depicted in \Cref{fig:Overview_module}. 

% Three crucial design will be introduced in this section: (1) Spatial-Temporal information learning module (SIL). (2) Knowledge spreading module (KSM). (3) Temporal confirmed pseudo-label (TPL).
% Our approach aims to transfer the spatial AU correlation knowledge from $F_{l}^{i}$ to $F_{u}^{i}$ and integrate the well-learned spatial information with semi-supervised $I^{i}$ for learning and fusing temporal knowledge and spatial knowledge. 

% To transfer and learn Spatial-Temporal knowledge in an effective way, our design concept conforms following basic rules: (1) The frame-level fully-supervised network F should learn relatively trustable spatial correlation knowledge independently. Thus, no unlabeled data should be fed to it for avoiding to confuse the model. (2) The video-level network V learn AU temporal correlation knowledge in a semi-supervised way. (2) Both spatial and temporal KD adopt the dual student design for bidirectional knowledge transferring. (3) The shifting operation of key frame should be balanced to avoid the generalization ability bias of sub-networks. (4) Any pseudo labels that violate temporal characteristics should be denied by the SSL module to to avoid the temporal perturbation or confirmation bias.

\subsubsection{Spatial-Temporal Information Learning (SIL)} SIL is an self-attention-based module for learning semi-supervised AU dependencies in both spatial and temporal domain. Transformers have been proved to be an effective method of learning informative temporal context in natural language processing. In order to address the limitation of applying conventional transformers on computer vision, ViT~\cite{Methodology_6} splits an image into fixed-sized patches to linearly embed them with position embedding. We extend the original ViT to learn the semantic relationship of AUs in both spatial-wise and temporal-wise. As shown in \Cref{fig:Overview_module}, the features from branch $A$ and branch $B$ are extracted by base models which are ResNet-18~\cite{Methodology_7} pre-trained on ImageNet~\cite{Methodology_4}. Model $S_{a}$ is designed for learning the spatial AU correlation. We firstly decouple the extracted global features $p_{s}$ into multiple branches by applying global-average-pooling. The collection of decoupled features is represented as $D_{s}= \left \{ f_{s}^{1}, f _{s}^{2}, \cdot \cdot \cdot , f _{s}^{u} \right \}$ where $u$ means the number of facial action units. Afterward, we assign $u$ 1D learnable positional embedding to be added with these features as AU-specific embedding. Then, the AU-specific embedding sequence is fed to a standard Transformer encoder for learning the spatial-wise AU correlation knowledge. Note that, the module does not need any extra classification token, for no obvious performance gains are observed by applying this design. Likewise, the temporal model $T_{b}$ in branch B also adopts the same model design. The collection of frame-specific features is depicted as $D_{b}= \left \{ p_{b}^{1}, p_{b}^{2}, \cdot \cdot \cdot , p_{b}^{n} \right \}$ where $n$ is the number of frames in each input sequence. The frame-specific features are extracted by the sub-networks in $S^{n}_{b}$ and added with the corresponding temporal embedding.

% , considering a one-to-one correspondence with labels of different AU embedding
% The difference is the non key-frames are supervised by pseudo-labels and there is no feature decoupling procedure.

% Unlike regular attention design, which ignores the task-specific dependencies, we design a simple but effective task-related module for modeling the facial action relationship. In addition, o
\subsubsection{Knowledge Spreading Module (KSM)}
KSM is the core module of the proposed Knowledge-Spreader. Inspired by FixMatch, our model also adopts a hybrid semi-supervised learning method by combining consistency regularization and pseudo-labeling. In this paper, we improve it to a video level method in the purpose of learning efficient AU dynamics. We approach KSM by several main steps: (1) train a group of Spatial Students $S_{b}$ which accommodate all the frames of input sequence, (2) let the length of input clip equals to the number of Spatial Students, (3) couple the input image in branch $A$ with the key frame of input video clip in branch $B$, (4) shift the location of key frame during training different batch of training samples, and (5) process distillation when Spatial Student gets the key frames, otherwise process pseudo-labeling. Furthermore, in terms of the distillation, we adopt two-level and bidirectional knowledge distillation design. The first level is from Spatial Teacher $S_{a}$ to Spatial Students $S_{b}$. For the MLP-based Spatial Students $S_{b}$, we apply data random augmentation and model dropout as the noise. We assume the knowledge learned by Spatial Teacher without noise is consistent with that learned by noisy Spatial Students. The second level is from Temporal Teacher $T_{b}$ to Temporal Student $S_{a}$. Instead of applying any artificial perturbation for temporal distillation, we utilize the original difference of input data for $T_{b}$ and $S_{a}$. We assume the spatial knowledge learned by $S_{a}$ should be consistent with the temporal knowledge learned by $T_{b}$. Thus, our model forces the Temporal Student $S_{a}$ to learn the temporal AU knowledge harder, even if it only gets the image-level inputs. Here, Model $S_{a}$ plays different roles in the two level distillation. Inspired by recent work~\cite{Methodology_3}, we adopt an online distillation method using KL divergence. The KL divergence loss in this work is used to minimize the probability distribution of student networks and their corresponding ensembles. The KL loss function of spatial knowledge distillation is defined as
\begin{equation}
L_{skd}=\frac{1}{b}\left (  \sum_{i=1}^{n}T^{2}KL\left ( p_{i}, q_{i} \right ) + \sum_{i=1}^{n}T^{2}KL\left ( w^{m}_{i}, q_{i} \right )\right )
\end{equation}
where $b$ is the batch size, $T$ is the temperature parameter. $p$ and $w$ denote the soften probability distribution calculated by the Spatial Teacher $S_{a}$ and one Spatial Student selected from $\left \{ S^{1}_{b}, S^{2}_{b}, ..., S^{n}_{b} \right \}$. The soft target $q$ is expressed as $q=softmax(z_{s}/T)$ where $z_{s}$ is performed by the mean pooling of the outputs from both $S_{a}$ and one selected $S_{b}$. $m$ is the key frame position used for controlling which Spatial Student is selected as the target of knowledge spreading. $m$ equals $B$ mod $n$, where $B$ is the $B$th batch of training samples, $n$ is the clip length. In this paper, temperature parameter is set as 1. The KL loss function of temporal knowledge distillation is defined as
\begin{equation}
L_{tkd}=\frac{1}{b}\left (  \sum_{i=1}^{n}T^{2}KL\left ( p_{i}, q_{i} \right ) + \sum_{i=1}^{n}T^{2}KL\left ( w_{i}, q_{i} \right )\right )
\end{equation}
$p$ and $w$ denote the soften probability distribution calculated by the Temporal Teacher $T_{b}$ and the Temporal Student $S_{a}$ respectively. Considering the data imbalance issues from skewing the training process and affect the performance of the model. We choose weighted BCE with logits as the multi-label classification loss of student networks and the function can be described as:
\begin{equation}
L_{bce}=-w [y log\sigma \left ( x \right ) +\left ( 1-y \right )log\left ( 1-\sigma \left (x\right ) \right ) ]
\end{equation} 
where $x$ refers to the output $O_{a}$ and $O_{b}$ for the loss function $L_s$ and $L_t$ in \Cref{eq:equation4}. $x$ also refers to the final ensemble output $O_{output}$ for $L_{bce}$ in \Cref{eq:equation1}. $\sigma(x)$ is the corresponding predicted probability. $y$ is the AU occurrence ground truth. The loss functions of spatial knowledge distillation and temporal knowledge distillation are expressed as:
\begin{equation}
\label{eq:equation4}
L_{s}=\sum_{i=1}^{z}L_{bce}^{i} + \alpha L_{skd}, L_{t}=\sum_{i=1}^{z}L_{bce}^{i} + \alpha L_{tkd}
\end{equation}
where $\alpha$ is the trade-off weight, $z$ is the number of student networks. Although branch $B$ contains multiple student networks $\left \{ S^{1}_{b}, S^{2}_{b}, ..., S^{n}_{b} \right \}$, only one of them is selected for calculating the loss function at the same time. Thus, $z$ equals 2. In this paper, $\alpha$ is set as 0.5.

Pseudo-label~\cite{Methodology_9} is a simple but efficient formulation of training models in a semi-supervised way. We deploy it to train the Spatial Students in $\left \{ S^{1}_{b}, S^{2}_{b}, ..., S^{n}_{b} \right \}$ when unlabeled frames are assigned as the input. Note that, the labeled data and unlabeled data are jointly trained. The loss function is defined as
\begin{equation}
L_{pdbce}=-w [\hat{y} log\sigma \left ( x \right ) +\left ( 1-\hat{y} \right )log\left ( 1-\sigma \left (x\right ) \right ) ]
\end{equation} 
where $\hat{y}$ denotes the pseudo labels of unlabeled frames by picking up the class which has the maximum predicted probability. The total loss function of semi-supervised learning can be denoted as
\begin{equation}
\label{eq:equation2}
L_{semi}= \sum_{i=1}^{n-1}L_{pdbce}^{i}
\end{equation}

\begin{figure}
\begin{center}
\includegraphics[width=0.5\linewidth]{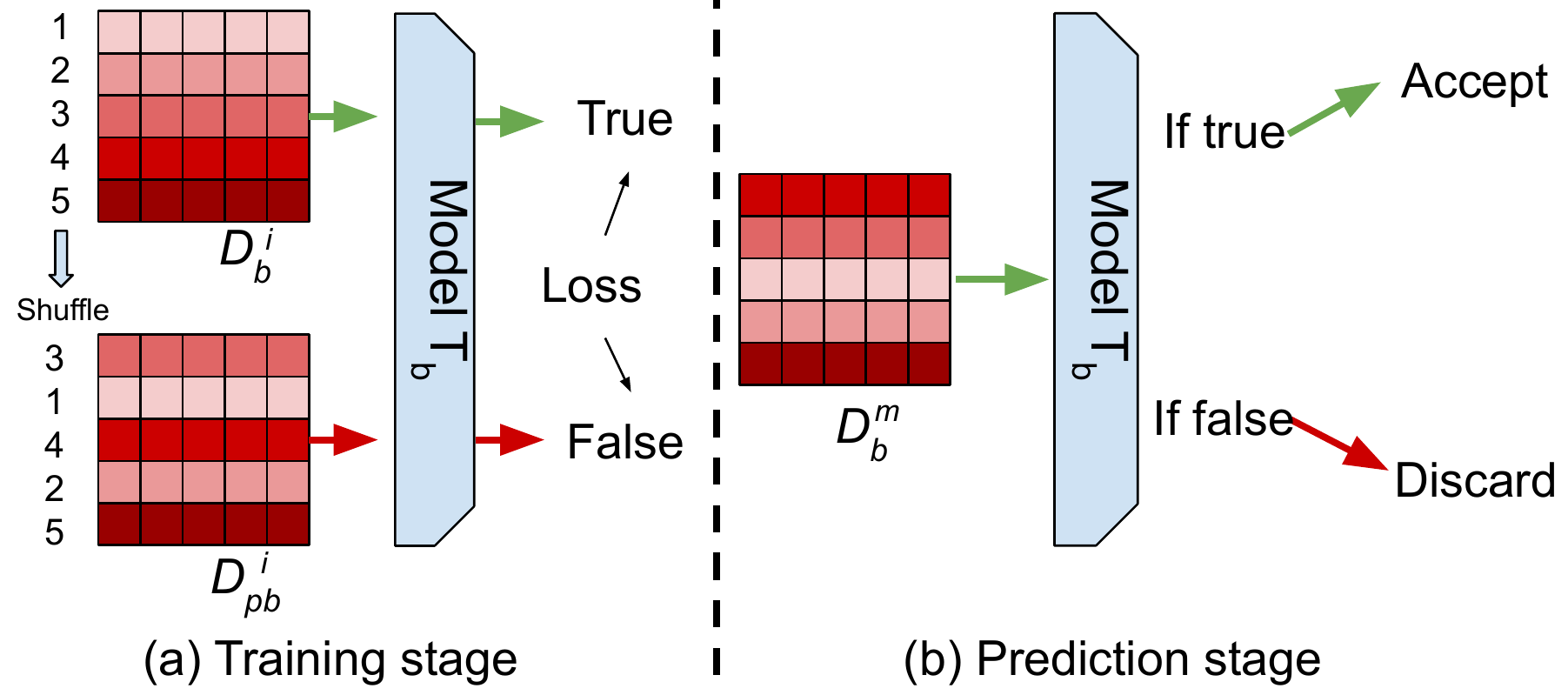}
\caption{Pipeline of the temporal confirmed pseudo-label. The module is based on self-supervised learning for a binary classification task. At training stage, it feeds $D^{i}_{b}$ (without shuffling) and $D^{i}_{pb}$ (shuffling $D^{i}_{b}$ along the timeline of the features) into the model. At the prediction stage, we let the feature $D^{m}_{b}$ (without shuffling) as the input. If it is classified as false, the corresponding pseudo labels will be discarded and vice versa. The training stage starts from the first epoch. By observation, we find the simple binary classifier takes only one or two epochs to convergence. Thus, we set the third epoch as the flag to start the prediction stage.}
\label{fig:pseudo-label}
\end{center}
\end{figure}

\subsubsection{Temporal Confirmed Pseudo-label (TPL)}
\label{sect:pd-confirmation} We exploit a simple self-supervised method for confirming confident pseudo labels by predicting if the features have any temporal perturbation. The smooth evolution of facial muscles leads that the facial appearance also moves gradually and smoothly over time. Thus, the pseudo labels generated by incorrect features contains anomalies in the temporal domain. Researchers~\cite{Methodology_5} utilize the self-supervised sequential perturbation to improve their model's robustness and generalization. Inspired by this work, we use the temporal feature shuffling to simulate the sequential perturbation and generate negative feature samples. Although a few video clips contain fully still frames which may diminish the contribution of temporal feature shuffling, they only occupy a small portion in the whole dataset. In addition, these samples do not contain any temporal information. Therefore, labeling them incorrectly does not affect the learning ability of the spatial model. The auxiliary task is defined as a binary classification task that is jointly trained with the AU classifier on Model $T_{b}$. $L_{ssl}$ denotes the loss of the binary classification task. Specifically, We first label the $i$th collection of sequential features $D^{i}_{b}$ as true while the shuffled feature collection $D^{i}_{pb}$ as false. Afterward, $D^{i}_{b}$ and $D^{i}_{pb}$ are sequentially fed into Model $T_{b}$. The loss $L_{ssl}$ encourages the model to classify them correctly. \Cref{fig:pseudo-label} illustrates the pipeline of the proposed module. As shown in \Cref{fig:confirmation}, the negative samples show the irregular pattern that AU occurs and disappears repeatedly in a short-term period. By detecting the temporal perturbation of the features and filtering out untrustable pseudo labels, TPL reliefs the confirmation bias issue existing in conventional pseudo-labeling methods.

\subsubsection{Overall loss function and algorithm}
The total loss is composed of the losses from previous sections, as follows:
\begin{equation}
\label{eq:equation1}
L_{total}= \lambda _{1}L_{s}+  w_{ramp} \left (  L_{bce}+ \lambda _{2}L_{t}+\lambda _{3}L_{ssl}+\lambda _{4}L_{semi} \right )
\end{equation}
where $w_{ramp}$ is a ramp-up function to make sure the semi-supervised learning and self-supervised learning converge relatively slowly compared with the fully-supervised task. It is inspired by~\cite{Methodology_10}. The function $\mu$, as a simple Gaussian curve function, is defined as:
\begin{equation}
w_{ramp} =exp\left ( -\omega (1-\frac{(x-\mu )^{2}}{\sigma ^{2}}) \right )
\end{equation}
where $x$ is the epoch number. In this paper, we set $\omega=2$, $\mu=0$, and $\sigma=5$. Here the $w_{ramp}$ is set as 1 after 5 epochs' warming-up for getting the optimal effect. The algorithm of KS is shown in \Cref{euclid}.

\begin{algorithm}
\scriptsize
\caption{Pseudocode of Knowledge-Spreader}\label{euclid}
\begin{algorithmic}[1]
\Require{The input frame $F_{l}^{i}$, the input clip $V^{i}$ and its frame number $N$, the position of key frame $m$. $B$ means the $B$th batch of training sample. Functions of the models in branch A: base model $b_{\theta }(x)$, Model $S_{a}$ $f_{\theta }(x)$. Functions of the models in branch B: base model $b_{\sigma }(x)$, sub-networks in $S^{n}_{b}$ with supervision $l_{\sigma }(x)$ and pseudo-labeling $k_{\sigma }(x)$, The AU detection classifier of Model $T_{b}$ $f_{\sigma }(x)$ and the binary classifier for self-supervised learning $g_{\sigma }(x)$.}{}
\For{each epoch \textit{E}}
\For{each mini-batch \textit{b}}
\State $O_{abase}^{i} \gets b_{\theta }(F_{l}^{i}); O_{bbase}^{(i,n)} \gets b_{\sigma }(V^{i})$
% \State $$
\State $O_{a}^{i} \gets f_{\theta }(O_{abase}^{i})$
\For{\texttt{ $q$ $=$ 1,...,$N$}}
\If {$q == B$ mod $N$} 
\State $O_{k}^{i} \gets l_{\sigma  }(O_{bbase}^{(i,q)})$
\State $\text{process spatial KD with } O_{a}^{i} \text{ and } O_{k}^{i}$
\Else
\State $\text{process pseudo-labeling with } O_{ps}^{(i, n-1)} \gets k_{\sigma  }(O_{bbase}^{(i,q)})$
\EndIf
\EndFor
\State $ O_{bti}^{i}\gets \text{Concatenate(}O_{ps}^{(i, n-1)}, O_{k}^{i})$
\State $ O_{b}^{i}\gets f_{\sigma  }( O_{bti}^{i})$
\State $ O_{pb}^{i}\gets f_{\sigma  }(\text{Shuffle}( O_{bti}^{i}))$
\State $ O_{ssl}^{i}\gets g_{\sigma  }( O_{b}^{i} = 0, O_{pb}^{i} = 1)$
\If {$E <= 2$ or $O_{ssl}^{i} == 1$} 
\State $\text{Let the weight of }L_{semi} \text{, } \lambda _{4} = 0$ 
\EndIf
\State $\text{process temporal KD with } O_{b}^{i} \text{ and } O_{a}^{i}$
\State $O^{i}_{output} \gets \text{MeanValue}(O_{b}^{i}, O_{a}^{i})$
\State $\text{Update } \theta \text{ and } \sigma\text{ via SGD of \Cref{eq:equation1} }$
\EndFor
\EndFor
\end{algorithmic}
\end{algorithm}

% \Require{The input frame for branch A $F_{l}^{i}$, the input clip for branch B $V^{i}$, $N$ is the frame number of $V^{i}$, $m$ is the position of the labeled frame in the clip. Functions of the models in branch A: base model is $b_{\theta }(x)$ , model $S_{a}$ is $f_{\theta }(x)$. Functions of the models in branch B: base model is $b_{\sigma }(x)$, sub-networks of model $S^{n}_{b}$ with supervision $l_{\sigma }(x)$ and pseudo-labeling $k_{\sigma }(x)$, model $T_{b}$ is $f_{\theta }(x)$. The function of getting the mean value of output is $f_{mean}(x_{1}, x_{2})$.}{}

%------------------------------------------------------------------------
\section{Experiments}

% \begin{table}[hbt]
% \caption{Overview of existing facial expression datasets and our new dataset}
% \label{tab:dataset}
% \centering
% \small
% \renewcommand{\arraystretch}{1.0}
% \begin{tabular}{c|c|c}
% \hline
% Name        & Subjects      & Modalities       \\ \hline
% DISFA       & 27            & Video   \\ 
% MMI         & 44            & Video     \\
% BP4D        & 41            & Video, 3D   \\ 
% BU-3DFE     & 100           & Image, 3D     \\ 
% BU-4DFE     & 101           & Video, 3D   \\ 
% BP4D+       & 140           & Video, 3D, Thermal, Physiology \\
% MMSE (Ours) & 233           & Video, 3D, Thermal, Physiology \\

% \hline

% \end{tabular}
% \end{table}

\subsection{Datasets} %%%%%%%%%%%%%%%%%% DAta Set
% \begin{figure}
% \begin{center}
% \includegraphics[width=0.9\linewidth]{Figures/dataset.pdf}
% \caption{A sample sequence from our MME. 2D texture image, 3D shaded model, 3D texture model, thermal image, and physiological signal (respiration rate, blood pressure, EDA, heart rate) and corresponding AU occurrence are shown from top to bottom.}
% \label{dataset}
% \end{center}
% \end{figure}

\textbf{BP4D} and \textbf{DISFA}: BP4D~\cite{Experiments_14} and DISFA~\cite{Experiments_15} are are widely used benchmark databases for dynamic AU detection. We followed the experimental setting of the previous work~\cite{Relatedworks_20} to evaluate our approach for a fair comparison.

% BP4D contains 41 subjects and around 140,000 frames captured under laboratory environments. There are 27 subjects and around 130,000 frames in the DISFA database. Following the previous settings~\cite{} of DISFA, the frames with intensities equal to or larger than 2 are regarded as AU occurrences while the rest are absent. We divide both datasets into subject-independent 3 folds and report the performance through cross-validation for a fair comparison with other algorithms.

              %%%%%% Table Datasets %%%%%%
\textbf{New MME}: The existing facial action datasets are limited in terms of subjects number, diversity, and metadata. Thanks to the existing available multi-modal datasets ~\cite{Experiments_14} ~\cite{Experiments_17}, we extend to develop a new larger-scale multi-modal emotion (MME) database, which consists of \textbf{233} participants (132 females and 101 males). The data is significantly expanded in terms of participants number as compared to the existing databases: DISFA (27 subjects)~\cite{Experiments_15}, MMI (44 subjects)~\cite{Experiments_16}, BP4D (41 subjects)~\cite{Experiments_14}, BP4D+ (140 subjects)~\cite{Experiments_17}. Following ethical principles, our data collection was approved by the institutional review board (IRB). Each subject signed an informed consent form. A professional performer/interviewer applied a procedure containing 10 seamlessly-integrated tasks as~\cite{Experiments_14} ~\cite{Experiments_17} that resulted in effective elicitation of spontaneous emotions. The dataset was well-synchronized and aligned with multi-modalities including 3D geometric facial model, 2D facial videos, thermal videos, and physiology data sequences (\textit{e.g.} heart rate, blood pressure, skin conductance (EDA), and respiration rate). Around 94,000 frames were well-annotated by three expert FACS coders for AU coding. More details are described in the supplemental material. The new database is ready for public and will be released to the research community by the time of the paper being published. 

% A sequential sample from our MME is shown in Figure \ref{dataset}. 

\subsection{Implementation Details}  
We process the image by cropping off redundant area which is not relevant to face recognition. Then the images are resized to $224 \times 224 \times 3$ ($H \times W \times C$) to fit the model. Each of the training images is randomly rotated, flipped horizontally, and with color jitters (saturation, contrast, and brightness) for data augmentation. We choose SGD as the optimizer with a learning rate of 0.01 for 50 epochs. The model was implemented with Pytorch framework. The hyper-parameters in \Cref{eq:equation1} are set as $\lambda _{1}=0.5$, $\lambda _{2}=0.5$, $\lambda _{3}=0.2$, and $\lambda _{4}=0.25$. We use video clips with 5 frames for learning the temporal information.
% The number $N$ of attention branches is set as 7 and 5 on BP4D and DISFA respectively. About how the channel number $N$ affects the recognition result, we observe the performance curve with different number $N$ in Fig.\ref{branch_number}. It shows the result improves roughly as the number increases until it reaches a certain threshold. We infer that a high number $N$ will not only cause the saturation of branch diversity but also generate overlapping attention maps that may dominate the model training and drive the learning process to omit valuable information from other branches.

\subsection{Model analysis}

\begin{figure*}[ht]
\begin{center}
\includegraphics[width=0.99\linewidth]{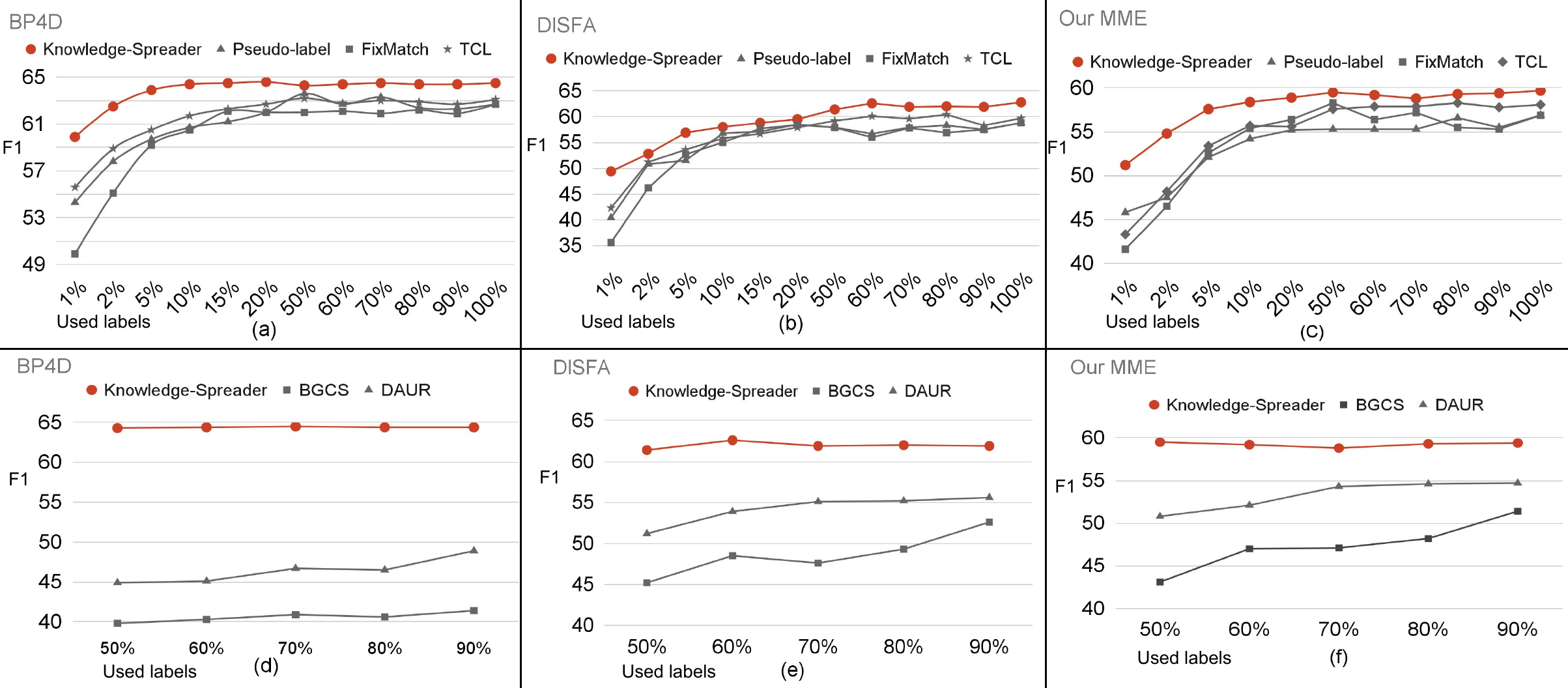}
\caption{Comparison with other advanced semi-supervised algorithms using different percentages of labels on BP4D, DISFA, and MME.}
\label{fig:semi_comparison}
\end{center}
\end{figure*}

\subsubsection{Comparison with semi-supervised methods}

\Cref{fig:semi_comparison} shows the performance comparison with semi-supervised methods from two areas (AU detection and general action recognition). We carefully investigated the existing works that adopt limited labels for AU detection. BGCS~\cite{Relatedworks_12} and DAUR~\cite{Relatedworks_11} are selected for comparison. \Cref{fig:semi_comparison} (d), (e), and (f) shows the proposed model achieves significant performance improvement. Considering our foundation model may have advantages in generalization ability, we can compare the performance trend. With the available labels decreasing (from 90\% to 50\%), Knowledge-Spreader shows no obvious performance attenuation. Note that some other semi-supervised models from~\cite{Introduction_8} ~\cite{Introduction_4} ~\cite{Experiment_5} ~\cite{Experiment_6} are not selected for comparison considering they use full annotation pools, extra data, or other jointly trained tasks. We further report the comparison results with some semi-supervised methods from general action recognition for a comprehensive evaluation, including Pseudo-label~\cite{Methodology_9}, FixMatch~\cite{Experiment_4}, and a video-level TCL~\cite{Experiment_1}. Compared with the conventional setting of previous AU works~\cite{Relatedworks_12} ~\cite{Relatedworks_11} ~\cite{Introduction_10} ~\cite{Introduction_8}, the percentages of available AU annotations are significantly reduced (1\%, 2\%, 5\%, 10\%, 20\%, 50\%, 60\%, 70\%, 80\%, 90\%, and 100\%) to explore where the limit of KS is. \Cref{fig:semi_comparison} (a), (b), and (c) shows prominent improvement of KS, especially when extremely limited annotations are available (1\%, 2\%, 5\% , 10\%, 15\% and 20\% on BP4D; 1\%, 2\%, and 5\% on DISFA; 1\%, 2\%, 5\% , 10\%, 15\% and 20\% on MME). More quantitative reports are shown in the supplementary.

\begin{table}[hbt]
\caption{Comparison with state-of-the-art methods using F1 score. The left table indicates other methods using 100\% labeled data. The right table indicates a baseline model (left) and the proposed KS (right) using different percentages of labels. Underlines indicate the best result of other methods. Bold numbers indicate KS surpasses others' best results. B and D indicates BP4D and DISFA.}
\label{tab:supervised_comparison}
\centering
\footnotesize
\renewcommand{\arraystretch}{1.0}
\resizebox{0.45\textwidth}{!}{
\begin{tabular}{L{2cm}|C{1.8cm}|C{1.3cm}|C{1.3cm}}
\hline
Model                           & Reference    & BP4D   & DISFA    \\ \hline
JAA~\cite{Experiments_8}         & ECCV'18      &60.0       &56.0           \\
DSIN~\cite{Experiments_9}       & ECCV'18       &58.9       &53.6           \\ 
LP~\cite{Experiments_10}        & CVPR'19       &61.0       &56.9      \\ 
ARL~\cite{Experiments_11}       & AC'19         &61.1       &58.7          \\ 
SRERL~\cite{Experiments_12}     & AAAI'19       &62.1       &55.9          \\
UGN~\cite{Experiments_13}       &AAAI'21        &63.3       &60.0\\
SEV~\cite{Introduction_2}       & CVPR'21       &63.9       &58.8          \\ 
HMP-PS~\cite{Relatedworks_7}     & CVPR'21      &63.4       &61.0        \\
FAUDT~\cite{Introduction_3}      & CVPR'21      &\underline{64.2}       &\underline{61.5}         \\
\hline
\end{tabular}
}
\resizebox{0.54\textwidth}{!}{
\begin{tabular}{L{2.1cm}|C{0.69cm}C{0.8cm}|L{1.6cm}|C{0.69cm}C{0.8cm}}
\hline
Model                           &B       &D          &Model              &B       &D  \\ 
\hline
EAC (1\%)~\cite{Experiment_7}   &43.8       &31.8           &KS (1\%)       &59.9           &49.4   \\ 
EAC (2\%)                       &48.7       &33.3           &KS (2\%)       &62.5           &52.8 \\ 
EAC (5\%)                       &52.2       &39.4           &KS (5\%)       &63.9           &56.9\\ 
EAC (10\%)                      &54.8       &43.9           &KS (10\%)      &\textbf{64.4}       &58 \\ 
EAC (15\%)                      &55.0       &45.1           &KS (15\%)      &\textbf{64.5}       &58.8 \\ 
EAC (50\%)                      &55.6       &48.0           &KS (50\%)      &\textbf{64.3}       &61.4\\ 
EAC (60\%)                      &55.9       &48.7           &KS (60\%)      &\textbf{64.4}       &\textbf{62.6}\\ 
EAC (100\%)                     &\underline{56.3} &\underline{51.2}  &KS (100\%)  &\textbf{64.5}       &\textbf{62.8}\\ 
\hline
\end{tabular}
}
\end{table}

% 15% and 100% need to run EAC

% EAC~\cite{Experiment_7}         & PAMI'18       &55.9       &48.5           \\ 

\begin{table*}[ht]          %%%%%%%%%%        TABLE BP$D             %%%%%%%%%%%
\caption{Comparison with state-of-the-art methods using F1 score in terms of individual AUs. The upper part is the F1 score on BP4D; The bottom part is the F1 score on DISFA. Bold numbers indicate the best performance.}
\label{tab:table1}
\footnotesize
\centering %
\renewcommand{\arraystretch}{1.0}
\resizebox{0.99\textwidth}{!}{
\begin{tabular}{l|c|*{12}{c}|c}
\hline
Model          & Used labels & AU1       & AU2       & AU4       & AU6       & AU7       & AU10      & AU12      & AU14      & AU15      & AU17      & AU23      & AU24      & Avg.\\
\hline
% DSIN            & 100\%     & 51.7      & 40.4      & 56.0      & 76.1      & 73.5      & 79.9      & 85.4      & 62.7      & 37.3      & 62.9      & 38.8      & 41.6      & 58.9\\
% LP              & 100\%     & 43.4      & 38.0      & 54.2      & 77.1      & 76.7      & 83.8      & 87.2      & 63.3      & 45.3      & 60.5      & 48.1      & 54.2      & 61.0\\
ARL             & 100\%     & 45.8      & 39.8      & 55.1      & 75.7      & 77.2      & 82.3      & 86.6      & 58.8      & 47.6      & 62.1      & 47.4      & 55.4      & 55.4\\
SRERL           & 100\%     & 46.9      & 45.3      & 55.6      & 77.1      & 78.4      & 83.5      & \B{87.6}  & 63.9      & \B{52.2}  & \B{63.9}  & 47.1      & 53.3      & 62.9\\ 
UGN             & 100\%     & 54.2      & 46.4      & 56.8      & 76.2      & 76.7      & 82.4      & 86.1      & 64.7      & 51.2      & 63.1      & 48.5      & 53.6      & 63.3\\
% SEV             & CVPR'21   & \B{58.2}  & \B{50.4}  & 58.3      & \B{81.9}  & 73.9      & \B{87.8}  & 87.5      & 61.6      & \B{52.6}  & 62.2      & 44.6      & 47.6      & 63.9\\
HMP-PS          & 100\%     & 53.1      & 46.1      & 56.0      & 76.5      & 76.9      & 82.1      & 86.4      & 64.8      & 51.5      & 63.0      & 49.9      & 54.5      & 63.4\\
FAUDT           & 100\%     & 51.7      & 49.3      & 61.0      & 77.8      & 79.5      & 82.9      & 86.3      & \B{67.6}  & 51.9      & 63.0      & 43.7      & 56.3      & 64.2\\
\hline
Our KS          & 15\%      & \B{58.7}  & \B{50.3}  & \B{62.0}  & \B{79.5}  & 75.4      & \B{84.9}  & 87.1      & 65.9      & 45.5      & 62.9      & 48.3      & 53.3      & \B{64.5}\\
Our KS          & 100\%     & 55.1      & 48.9      & 56.2      & 77.3      & \B{81.8}  & 83.3      & 86.4      & 62.6      & 51.9      & 61.3      & \B{51.0}  & \B{58.3}  & \B{64.5}\\
\hline
\end{tabular}
}
\resizebox{0.99\textwidth}{!}{
\begin{tabular}{l|c|*{8}{C{1.131cm}}|c}
\hline
Model          & Used labels & AU1       & AU2       & AU4       & AU6       & AU9       & AU12      & AU25      & AU26      & Avg.\\
\hline
% DSIN            & 100\%     & 42.4      & 39.0      & 68.4      & 28.6      & 46.8      & 70.8      & 90.4      & 42.2      & 53.6\\
% LP              & 100\%     & 29.9      & 24.7      & 72.7      & 46.8      & 49.6      & 72.9      & 93.8      & 65.0      & 56.9\\
ARL             & 100\%     & 43.9      & 42.1      & 63.6      & 41.8      & 40.0      & \B{76.2}  & \B{95.2}  & 66.8      & 58.7\\
SRERL           & 100\%     & 45.7      & 47.8      & 59.6      & 47.1      & 45.6      & 73.5      & 84.3      & 43.6      & 55.9\\ 
UGN             & 100\%     & 43.3      & 48.1      & 63.4      & 49.5      & 48.2      & 72.9      & 90.8      & 59.0      & 60.0\\
% SEV             & 100\%     & \B{55.3}  & 53.1      & 61.5      & 53.6      & 38.2      & 71.6      & \B{95.7}  & 41.5      & 58.8\\
HMP-PS          & 100\%     & 38.0      & 45.9      & 65.2      & 50.9      & \B{50.8}  & 76.0      & 93.3      & \B{67.6}  & 61.0\\
FAUDT           & 100\%     & 46.1      & 48.6      & \B{72.8 } & \B{56.7}  & 50.0      & 72.1      & 90.8      & 55.4      & 61.5\\
\hline
Our KS          & 15\%      & 41.7      & 53.5      & 69.7      & 41.3      & 46.2      & 72.0      & 92.3      & 54.0      & 58.8\\
Our KS          & 100\%     & \B{53.8}  & \B{59.9}  & 69.2      & 54.2      & \B{50.8 } & 75.8      & 92.2      & 46.8      & \B{62.8}\\
\hline
\end{tabular}
}

\end{table*}

\subsubsection{Comparison with supervised methods}
We report the results under two training setups as~\cite{Experiments_18}: (1) Compare KS against the fully-supervised state-of-the-art methods with 100\% labeled data. (2) Compare KS against a supervised counterpart under different training label ratios. As shown in \Cref{tab:supervised_comparison}, a collection of recent and strong benchmark algorithms are selected for better evaluation. Knowledge-Spreader outperforms all other advances using only 10\% labels on BP4D and 60\% labels on DISFA. KS still performs competitively using only 2\% labels on BP4D and 5\% labels on DISFA. In addition, the experiment conducted with 100\% labels shows the effeteness of Knowledge-Spreader in a supervised manner, where the main contribution comes from Spatial-Temporal information learning module. Especially, it shows that KS surpasses the best benchmark (FAUDT) by 1.3 f1-score on DISFA. \Cref{tab:table1} further shows the comparison results in terms of individual AUs. The proposed method performs best on 8 out of 12 AUs on BP4D and 3 out of 8 AUs on DISFA.

% , including EAC~\cite{Experiment_7}, JAA~\cite{Experiments_8}, DSIN~\cite{Experiments_9}, LP~\cite{Experiments_10}, ARL~\cite{Experiments_11}, SRERL~\cite{Experiments_12}, SEV~\cite{Introduction_2}, HMP-PS~\cite{Relatedworks_7}, and FAUDT~\cite{Introduction_3}, 

\subsection{Data Structure Analysis}
\begin{figure}
\begin{center}
\includegraphics[width=0.98\linewidth]{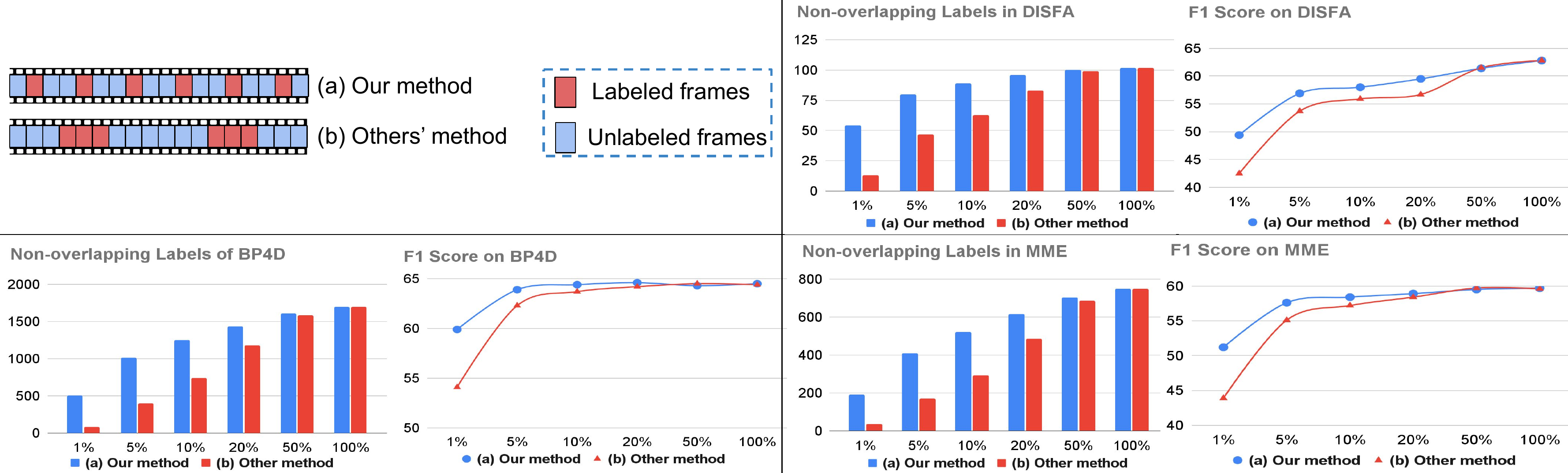}
\caption{Evaluation of the label sampling methods. The left top (a) indicates our strategy, and (b) indicates the conventional method. The left bottom, right top, and right bottom show the quantitative statistics of non-overlapping labels and the performance comparison of KS using different methods on BP4D, DISFA, and MME. X-axis refers to the percentage of used labels.}
\label{fig:data_structure}
\end{center}
\end{figure}

% The continuous annotations of video with high frame rate can provide rich and consistent time series information. On the other hand, it also inevitably contain unnecessary redundant annotations. 
Through the statistical experiments, we find the non-overlapping AU annotations (different AU combinations) account for only a small proportion of the overall frames number (1693 out of 140,000 frames in BP4D, 102 out of 130,000 frames in DISFA, 748 out of 94,000 frames in MME). A large number of similar labels and data densely exist across adjacent frames. It reveals that why using only a few \textbf{sparsely sampled} clips and annotations can achieve competitive or even better performance, which is consistent with the ``less is better'' principle from~\cite{Relatedworks_17}. Different from existing video-level semi-supervised works~\cite{Experiment_2} ~\cite{Methodology_5} ~\cite{Experiment_1} ~\cite{Experiment_3} that adopt continuous annotations, we sparsely sample the annotations and allocate only one annotation for every $k$ frames. \Cref{fig:data_structure} demonstrates that applying our method can reserve more non-overlapping AU annotations than conventional approaches with using the same number of annotations. Consequently, by utilizing abundant non-overlapping AU annotations, semi-supervised models testing on BP4D and MME is relatively more robust to resist the interference caused by missing annotations.

% This is one of the reason why BP4D can still achieve excellent performance with extremely limited labels.

\subsection{Ablation study}
\begin{table}\renewcommand\arraystretch{1}
\caption{ Ablation study on BP4D, DISFA, and MME using F1 score. Bold numbers indicate the best performance.}
\begin{center}
\footnotesize
\resizebox{0.99\textwidth}{!}{
\begin{tabular}{lcccccc} \toprule
    {Modules} & {BP4D (5\%)} & {DISFA (5\%)} & {MME (5\%)} & {BP4D (50\%)} & {DISFA (50\%)} & {MME (50\%)} \\ \midrule
     {Knowledge-Spreader}           & \textbf{63.9}   & \textbf{56.9} &  \textbf{57.6}  & \textbf{64.3}   & \textbf{61.4} &  \textbf{59.5}  \\ \midrule
     {KSM+TPL}                      & 61.6   & 54.8 &  55.5  & 62.8  & 60.2 &    58.1  \\
    {SIL+TPL}                       & 61.7   & 54.6 &  55.8  & 62.5  & 60.1 &   57.7  \\
    {SIL+KSM}                       & 62.6  & 55.1 &   55.2  & 63.7   & 60.6 & 58.4  \\
    \bottomrule
\end{tabular}
}
\label{tab:ablation_study}
\end{center}
\end{table}
% \Cref{tab:ablation_study} demonstrates the results of the ablation study for investigating the effectiveness of each module in KS.
% \midrule
%      {Modules} & {BP4D (50\%)} & {DISFA (50\%)} & {MME (50\%)} \\ \midrule
%       {Knowledge-Spreader}            & \textbf{64.3}   & \textbf{61.4} &  \textbf{59.5}    \\ \midrule
%     {KSM+TPL}          & 62.8  & 60.2 &    58.1 \\
%     {SIL+TPL}         & 62.5  & 60.1 &   57.7 \\
%     {SIL+KSM}              & 63.7   & 60.6 & 58.4 \\

\begin{figure}[ht]
\begin{center}
\includegraphics[width=0.5\linewidth]{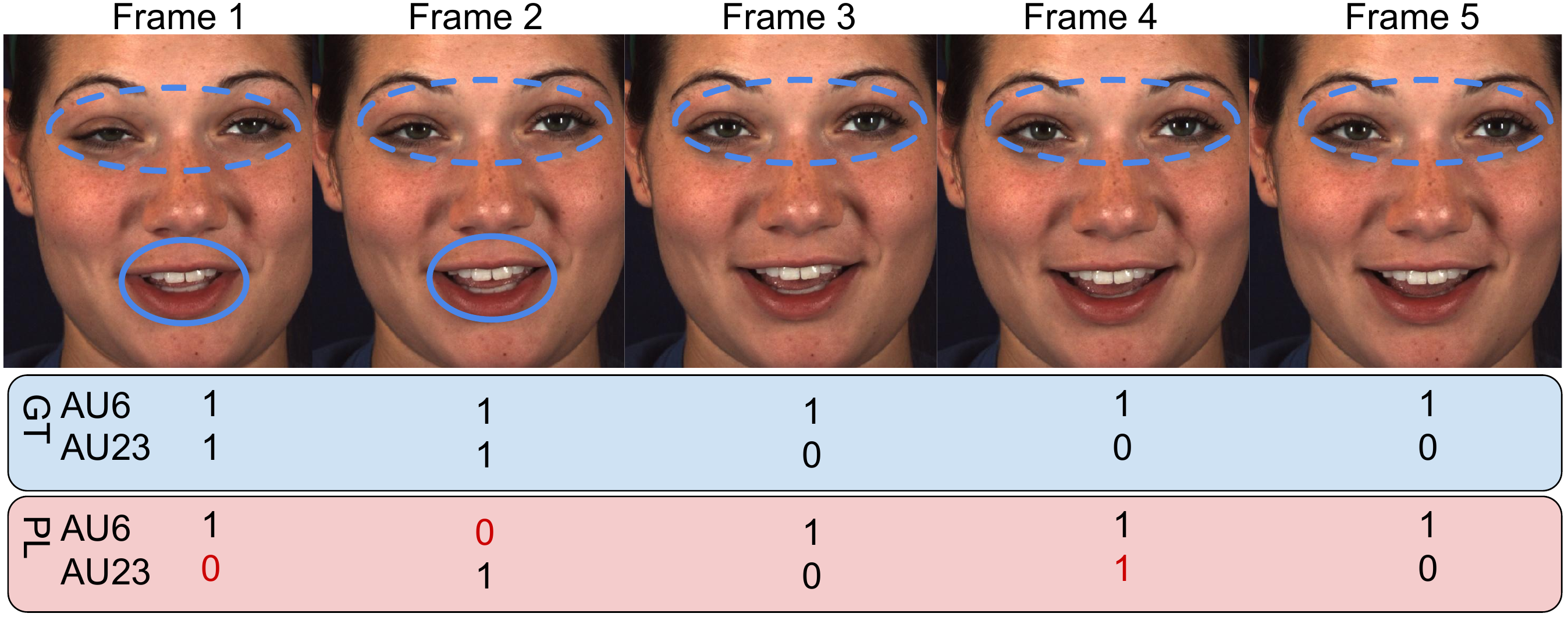}
\caption{Pseudo labels discarded by TPL. The dotted and solid circles indicate the occurrence of AU6 and AU23. ``GT'' and ``PL'' indicate the ground-truth and the pseudo labels. Red number means incorrect pseudo labels. The ground-truth labels usually present constant pattern with no change (GT of AU6 in this figure) or simple changes (GT of AU 23 in this figure), while the incorrect pseudo labels (PL of AU6 and AU23) with temporal perturbation show the pattern that AU occurs and disappears repeatedly in a short-term period.}
\label{fig:confirmation}
\end{center}
\end{figure}

\textbf{Effect of the Spatial-Temporal information learning module (SIL)}: 
We replace the Transformer-based SIL design with a vanilla MPL module to learn the spatial information and integrate them for learning temporal cues for a fair comparison. As shown in \Cref{tab:ablation_study}, the simple design leads to obvious performance degradation of KS due to being incapable of learning the AUs correlation knowledge.

\textbf{Effect of the knowledge spreading module (KSM)}:
We keep the design of key frame shifting operation, while remove the spatial knowledge and temporal knowledge distillation by setting the corresponding loss $L_{s}$ and $L_{t}$ to be unavailable. As a core module of Knowledge-Spreader, it plays an important role in fully spreading the AU semantic knowledge. There is an obvious performance gap between the model with and without KSM. For instance, without this module, the F1 score decreases 2.2, 2.3, and 1.8 on three databases with 5\% labeled data. Similarly, it decreases 1.8, 1.3, and 1.8 with 50\% labels. It demonstrates that the performance degradation of the model without KSM becomes more serious as the affordable labels decrease.

\textbf{Effect of the temporal confirmed pseudo-label (TPL)}:
As shown in \Cref{tab:ablation_study}, with the assistant of TPL, the performance of KS is improved under different circumstances. In addition, we compare the accuracy of pseudo labels generated by TPL and naïve pseudo-label design on BP4D using 10\% labels. We get 76.35\% accuracy with TPL and 73.36\% with the original pseudo-label, which shows the TPL achieves prominent improvement by discarding the incorrect pseudo labels with temporal perturbation. A sample is given in \Cref{fig:confirmation} to illustrate how TPL senses and processes the incorrect pseudo labels. All the above ablation studies prove that our complete model performs the best by integrating the three key components.

% \begin{table}[hbt]
% \caption{Cross-database evaluation using F1 score.}
% \label{tab:MME_dataset}
% \centering
% \footnotesize
% \renewcommand{\arraystretch}{1.2}
% \begin{tabular}{lccccccc}
% \hline
% Dataset & AU1    & AU6   & AU7   &AU10   & AU12 & AU17 &Avg   \\ \hline
% BP4D+   & 27.7   & 83.0  & 86.7  & 89.1  & 82.6 & 39.6 &68.1         \\
% MME     & 45.7   & 76.6  & 84.8  & 86.6  & 88.6 & 36.7 &69.8           \\ 
% \hline
% \end{tabular}
% \end{table}

% \subsection{Evaluation on MME}
% To further evaluate the annotation quality of MME, we conduct a cross-database experiment by training KS with the data from BP4D and testing it on BP4D+ and the new MME. The result can be seen in \Cref{tab:MME_dataset}.

%%%%%%%%%%%%%%%%%%%%%%%%%%%%%%%%%%%%%%%%%%%%%%%%%%%%%%%%%%%%%%%%%%%%%%%%%%%%%%%%
\section{Conclusion}
In this paper, we propose a new unified Knowledge-Spreader architecture to learn the interactive Spatial-Temporal correlation knowledge with sparsely labeled videos by incorporating dynamic knowledge distillation, pseudo-labeling, and AU semantic encoding modules. Results show that the proposed model using extremely limited annotations achieves superior performance than existing methods. Comprehensive studies have demonstrated the key factor that leads to the success of Knowledge-Spreader in terms of model design and data structure, in hope of inspiring future works. In addition, a large-scale dataset for spontaneous and dynamic facial action analysis is introduced to alleviate the scarcity issue of AU annotation and subject samples. In the future, we plan to explore the application of Knowledge-Spreader on multi-modal recognition tasks. This material is based on the work supported by the US National Science Foundation.

%  Comprehensive studies have demonstrated the key factor that leads to the success of Knowledge-Spreader in terms of model design and data structure, in hope of inspiring future works. This material is based on the work supported by the National Science Foundation (grant \# is removed for double blind review).

\section{Supplementary of Model Design}

\subsection{Details of the Spatial-Temporal Information Learning Module}

\begin{figure}
\begin{center}
\includegraphics[width=1.0\linewidth]{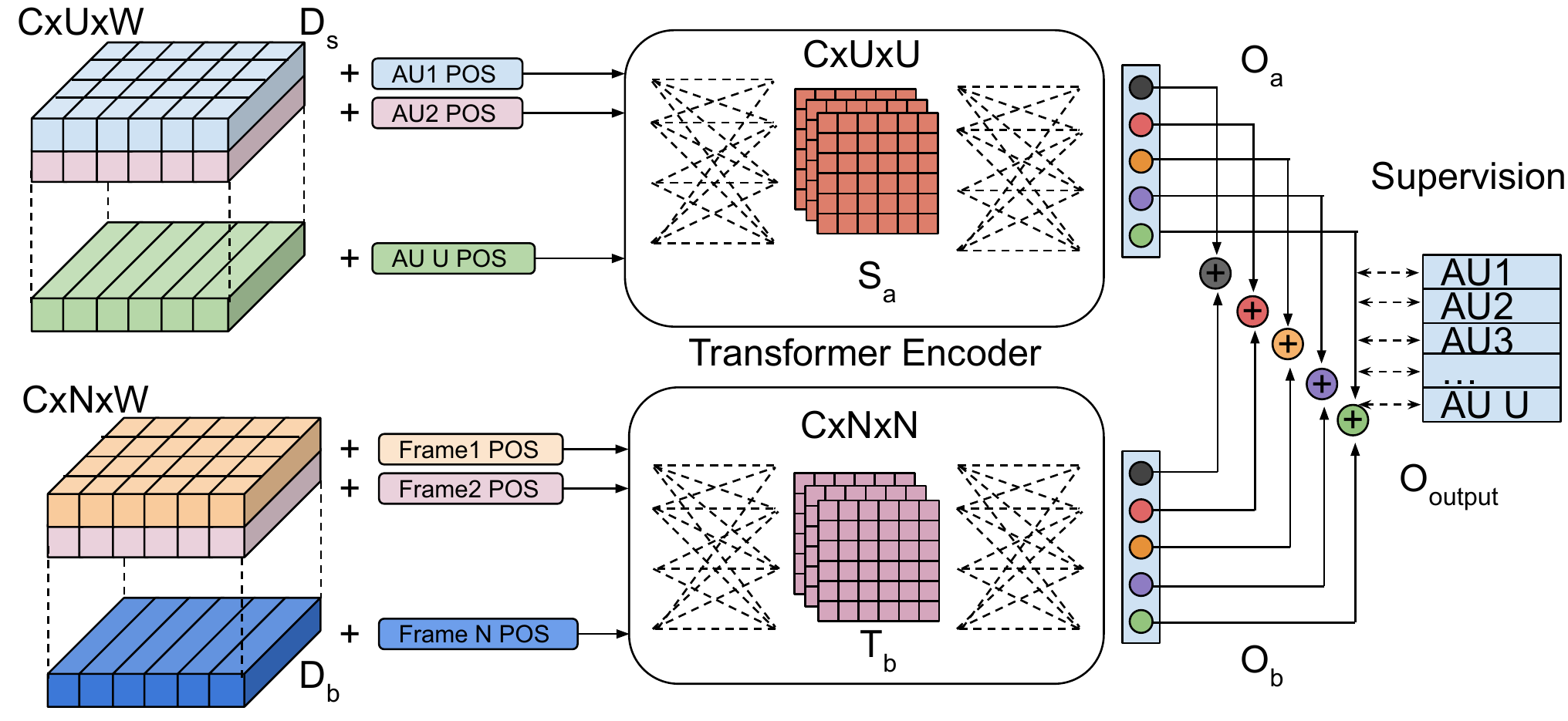}
\caption{Illustration of transformer-based model $S_{a}$ and model $T_{b}$ for learning spatial and temporal AU correlation.}
\label{attention-module}
\end{center}
\end{figure}

The Spatial Teacher (or Temporal Student) $S_{a}$ and the Temporal Teacher $T_{b}$, in \Cref{attention-module}, are designed for modeling the relationships between AUs in both the intra-frame and inter-frame levels. In Spatial-Temporal Information Learning (SIL) module, we first assign $U$ 1D learnable AU positional embedding to be added with input feature set $D_{s}$ as AU-specific embedding, where $U$ means the number of facial action units. For $u$th AU-specific features, a ViT transformer encoder $S_{a}$ generate a set of query, key, and value tensors ($Q_{u}, K_{u}, V_{u}$), each with dimension $\mathbb{R}^{C\times W}$. Afterwards, a $C \times U \times U$ learnable attention matrix is obtained as follows:
\begin{equation}
A^{u}=softmax\left ( \frac{Q_{u}K_{u}^{U}}{\sqrt{H}} \right )
\end{equation}
Similarly, we assign $N$ 1D learnable frame positional embedding to be added with input feature set $D_{b}$ as frame-specific embedding, where $N$ means the length of the input sequence. For $n$th frame in a sequence, another set of query, key, and value tensors ($Q_{n}, K_{n}, V_{n}$) are generated with the dimension $\mathbb{R}^{C\times W}$. A $C \times N \times N$ attention matrix is obtained as follows:
\begin{equation}
A^{n}=softmax\left ( \frac{Q_{n}K_{n}^{N}}{\sqrt{H}} \right )
\end{equation}
Through a series of dimensional changes, model $S_{a}$ and $T_{b}$ outputs $U$ dimensional tensors $O_{a}$ and $O_{b}$ respectively. Finally, we take the mean ensemble output $O_{output}$ of $O_{a}$ and $O_{b}$ for multi-label prediction.

Besides, we design $N$ small-sized MLP-based models for the Spatial Students $S_{b}$. The dimension of each output feature $p_{b}$ for Spatial Student is $C \times 1 \times W$.

\subsection{Q\&A for Model Design and Experiments}
In this section, we explain some doubts of the model design and experiments.

\textbf{Why the knowledge spreading operation matters for semi-supervised learning?} How to maximize the use of a small but credible learned knowledge to spread into a large amount of unknown data is the core problem solved by KS. The dynamic knowledge spread with two-level distillation, which can also be seen as a two-level lever, can allow KS to transform a few spatial knowledge into a large amount of spatial knowledge, and then form more temporal knowledge. Compared with static knowledge distillation, dynamic knowledge spread contains multiple level of leverage to infer more out-of-distribution knowledge with the fewest labels.

\textbf{What's the difference between general Spatial-Temporal information and our Spatial-Temporal AU correlation knowledge?} Two attention matrices, in \Cref{attention-module}, contains the relevance of each atomic AU class and frame-level class. By modeling the spatial and temporal AU dependency in attention matrices, the video-level and frame-level AU co-occurrence and mutual exclusive relation is refined and learned to improve the the general Spatial-Temporal information.

\textbf{Why do we choose transformer?} First, self-attention based methods (i.e., JAAnet~\cite{Experiments_8}) have been proved to be very effective in learning AUs semantic relation. Second, the residual connection, multi-head attention, and positional embedding designs make it an efficient tool to learn long-range temporal cues and more discriminative representations. Third, our experiments show that transformer performs better than other popular models such as regular GCNs.

\textbf{Why do we need both consistency regularization and pseudo-labeling?} The consistency regularization is used for building two level knowledge distillation. It distills the fully modeled spatial AU relationship knowledge from branch A to branch B, and distill the temporal AU correlation knowledge from branch B to branch A. However, without pseudo-labeling, some of the Spatial Students stay idle and the unlabeled data is not fully utilized, making the training process less efficient. By combining consistency regularization and pseudo-labeling, KS can accelerate the speed and effect of knowledge distillation and spread.

\textbf{Why are Spatial Student models based on MLP instead of transformer?} In most cases, KS need feed Spatial Student models with unlabeled data, which means the results derived by Spatial Students is less credible. Thus, a weak design of Spatial Student is necessary. In additional, different design of Spatial Student amd Spatial Teacher can be recognized as a noise or a model-wise perturbation for better applying the consistency regularization.

\textbf{Why do we need the temporal constrain for pseudo-labeling?} First, automatic AU detection, as a multi-label task, faces some difficulties in selecting and retaining the pseudo labels when only partial labels are with high confidence score. Compared with multi-class tasks, the one-hot format of multi-labels makes it hard to decide if the pseudo labels are confident in a holistic way or a local way. Setting the temporal constrain makes it easier to filter the cases which are not consist with the regular pattern of facial action movements. Second, the temporal label smoothness is also a soft constrains from human knowledge for better generalizing out-of-distribution AU data.

\textbf{How to generate pseudo labels?} We pick up the class which has maximum predicted probability for each binary class of unlabeled samples.
% \begin{equation}
% \hat{y}_{i}=\left\{\begin{matrix}
% 1 \text{ if } i ==\text{argmax}i^{'}f_{i^{'}(x)} \\
% 0 \text{ }\text{ }\text{ }\text{ }\text{ }\text{ }\text{ }\text{ }\text{ }\text{ }\text{ }\text{ }\text{ }\text{ }\text{ }\text{ }\text{ }\text{otherwise}
% \end{matrix}\right.
% \end{equation} 

\textbf{Can KS be trained with video clips without any labels? If yes, how?} Yes, it can. When feeding KS with the unlabeled video clips, we only update our model with the loss of pseudo-labeling and Temporal Teacher. It worth noting that the unlabeled video clips are not allowed to use before the model training is stable (10 epochs in this project). Otherwise learning unreliable knowledge first will lead the model to the error-prone issue.

\textbf{Why using 20\% labels achieves nearly the same performance as using 100\% on BP4D?}  This is due to the large portion of overlapped annotations, using 20\% labels with sparsely sampled annotations makes the performance reach the “saturation” quickly, hence there is no significant performance gain after 20\% towards the use of 100\% labels. Our finding complies with the “less is better” principle confirmed by the other existing works~\cite{Relatedworks_17}.

\textbf{How about applying different sequential perturbation for the pseudo label confirmation module?} The fact is that using or mixing certain perturbations (e.g., temporal feature shift, random mask, and flip) does not bring obvious performance improvement. We speculate that other perturbations can not model the temporal fluctuations caused by incorrect pseudo labels well. Applying inappropriate or excessive perturbation operation can even degrade the performance.

\subsection{Inference Strategy}
We adopt the inductive learning manner at testing stage. A random key frame number is assigned to each batch of input video clips. Given a video clip, both branch A and branch B are used to extract the feature $O_{output}$ for prediction.

\subsection{Additional Implementation Details}

All training images in the same video clips are randomly rotated (-45 to 45 degrees), flipped horizontally (50\% possibility), and with color jitters (saturation, contrast, and brightness) simultaneously. The detailed specification of Knowledge-Spreader is shown in the original code (model designing part). The complete code will be released to the research community by the time of the paper being published. We choose 5 as the frame length of each input video clip for the optimal time-and-accuracy trading-off. The analysis of the hyper-parameters can be seen in \Cref{hyper-parameter}. We implement our Knowledge-Spreader (KS) with the Pytorch framework and perform training and testing on the NVIDIA GeForce 2080Ti GPU.

\subsection{Additional Quantitative Evaluation}

The quantitative results with different label ratios are shown in \Cref{tab:supervised_comparison} for reference. It corresponds to Figure 4 in the original paper. In addition, due to the page limitation, only partial comparison results with supervised methods in terms of individual AU are shown in Table 2 of the original paper. \Cref{tab:table1} shows the complete comparison results.

\begin{table}[hbt]
\caption{Quantitative comparison with semi-supervised methods using F1 score. Underlines indicate the best results of individual models.}
\label{tab:supervised_comparison}
\centering
\footnotesize
\renewcommand{\arraystretch}{1.0}
\resizebox{0.47\textwidth}{!}{
\begin{tabular}{L{3cm}|C{1.3cm}|C{1.3cm}|C{1.3cm}}
\hline
Model                           & BP4D    & DISFA   & MME    \\ \hline
Pseudo-label (1\%)              & 54.3      &40.4       &45.8           \\
Pseudo-label (2\%)              & 57.8       &50.8       &47.5           \\ 
Pseudo-label (5\%)              & 59.7       &51.5       &52.1      \\ 
Pseudo-label (10\%)              & 60.7     &56.8       &54.2          \\ 
Pseudo-label (15\%)              & 61.2       &57.1       &54.9          \\
Pseudo-label (20\%)              & 62        &58.5       &55.2   \\
Pseudo-label (50\%)              & \underline{63.6}       &57.9       &55.3          \\ 
Pseudo-label (60\%)              & 62.7      &56.7       &55.3        \\
Pseudo-label (70\%)              & 63.3      &57.9       &55.3\\
Pseudo-label (80\%)              & 62.4      &58.3       &56.6        \\
Pseudo-label (90\%)              & 62.3      &57.5       &55.5        \\
Pseudo-label (100\%)              & 62.7      &\underline{58.8}       &\underline{56.9}        \\
\hline
\end{tabular}
}
\resizebox{0.47\textwidth}{!}{
\begin{tabular}{L{3cm}|C{1.3cm}|C{1.3cm}|C{1.3cm}}
\hline
Model                           & BP4D    & DISFA   & MME    \\ \hline
FixMatch (1\%)              & 49.9      &35.6       &41.6           \\
FixMatch (2\%)              & 55.1       &46.2       &46.5           \\ 
FixMatch (5\%)              & 59.2       &52.7       &52.6      \\ 
FixMatch (10\%)              & 60.5         &55       &55.4          \\ 
FixMatch (15\%)              & 62.1       &57.7       &55.6          \\
FixMatch (20\%)              &62        &58.4       &56.4      \\
FixMatch (50\%)              & 62       &57.9       &\underline{58.3}          \\ 
FixMatch (60\%)              & 62.1      &56       &56.4        \\
FixMatch (70\%)              & 61.9      &57.8     &57.2        \\
FixMatch (80\%)              & 62.2      &56.9       &55.5        \\
FixMatch (90\%)              & 61.9      &57.5       &55.3        \\
FixMatch (100\%)              & \underline{62.7}      &\underline{58.8}       &56.9        \\
\hline
\end{tabular}
}
\resizebox{0.47\textwidth}{!}{
\begin{tabular}{L{2.40cm}|C{1.3cm}|C{1.3cm}|C{1.3cm}}
\hline
Model                           & BP4D    & DISFA   & MME    \\ \hline
TCL (1\%)              &55.6       &42.3       &43.3           \\
TCL (2\%)              &58.9        &51.2       &48.2           \\ 
TCL (5\%)              &60.5       &53.6       &53.4      \\ 
TCL (10\%)              &61.7         &55.8       &55.7          \\ 
TCL (15\%)              &62.3       &56.7       &56.2          \\
TCL (20\%)              &62.7        &57.9       &55.6\\
TCL (50\%)              &\underline{63.2}       &59.2       &57.6          \\ 
TCL (60\%)              &62.8      &60.1       &57.9        \\
TCL (70\%)              &63.0      &59.6       &57.9\\
TCL (80\%)              &62.9      &\underline{60.4}      &\underline{58.3}        \\
TCL (90\%)              &62.7      &58.3       &57.8        \\
TCL (100\%)              &63.1      &59.7       &58.1        \\
\hline
\end{tabular}
}
\resizebox{0.47\textwidth}{!}{
\begin{tabular}{L{2.40cm}|C{1.3cm}|C{1.3cm}|C{1.3cm}}
\hline
Model                           & BP4D    & DISFA   & MME    \\ \hline
Our KS (1\%)              & 59.9      &64.9       &51.2           \\
Our KS (2\%)              & 62.5       &52.8       &54.8           \\ 
Our KS (5\%)              & 63.9       &56.9       &57.6      \\ 
Our KS (10\%)              & 64.4         &58       &58.4          \\ 
Our KS (15\%)              & 64.5       &58.8       &58.7          \\
Our KS (20\%)              & \underline{64.6}        &59.5       &58.9\\
Our KS (50\%)              & 64.3       &61.4       &59.5          \\ 
Our KS (60\%)              & 64.4      &62.6       &59.2        \\
Our KS (70\%)              & 64.5      &61.9       &58.8\\
Our KS (80\%)              & 64.4      &62       &59.3        \\
Our KS (90\%)              & 64.4      &61.9       &59.4        \\
Our KS (100\%)              & 64.5      &\underline{62.8}       &\underline{59.7}        \\
\hline
\end{tabular}
}
\end{table}

\begin{table*}[ht]
\caption{Comparison with state-of-the-art methods using F1 score in terms of individual AUs. The upper part is the F1 score on BP4D; The bottom part is the F1 score on DISFA. Bold numbers indicate the best performance.}
\label{tab:table1}
\scriptsize
\centering %
\renewcommand{\arraystretch}{1.0}
\begin{tabular}{l|c|*{12}{c}|c}
\hline
Model          & Used labels & AU1       & AU2       & AU4       & AU6       & AU7       & AU10      & AU12      & AU14      & AU15      & AU17      & AU23      & AU24      & Avg.\\
\hline
DSIN            & 100\%     & 51.7      & 40.4      & 56.0      & 76.1      & 73.5      & 79.9      & 85.4      & 62.7      & 37.3      & 62.9      & 38.8      & 41.6      & 58.9\\
JAA            & 100\%     & 47.2      & 44.0      & 54.9      & 77.5      & 74.6      & 84.0      & 86.9      & 61.9      & 43.6      & 60.3      & 42.7      & 41.9      & 60.0\\
LP              & 100\%     & 43.4      & 38.0      & 54.2      & 77.1      & 76.7      & 83.8      & 87.2      & 63.3      & 45.3      & 60.5      & 48.1      & 54.2      & 61.0\\
ARL             & 100\%     & 45.8      & 39.8      & 55.1      & 75.7      & 77.2      & 82.3      & 86.6      & 58.8      & 47.6      & 62.1      & 47.4      & 55.4      & 55.4\\
SRERL           & 100\%     & 46.9      & 45.3      & 55.6      & 77.1      & 78.4      & 83.5      & \B{87.6}  & 63.9      & \B{52.2}  & \B{63.9}  & 47.1      & 53.3      & 62.9\\ 
UGN             & 100\%     & 54.2      & 46.4      & 56.8      & 76.2      & 76.7      & 82.4      & 86.1      & 64.7      & 51.2      & 63.1      & 48.5      & 53.6      & 63.3\\
% SEV             & CVPR'21   & \B{58.2}  & \B{50.4}  & 58.3      & \B{81.9}  & 73.9      & \B{87.8}  & 87.5      & 61.6      & \B{52.6}  & 62.2      & 44.6      & 47.6      & 63.9\\
HMP-PS          & 100\%     & 53.1      & 46.1      & 56.0      & 76.5      & 76.9      & 82.1      & 86.4      & 64.8      & 51.5      & 63.0      & 49.9      & 54.5      & 63.4\\
FAUDT           & 100\%     & 51.7      & 49.3      & 61.0      & 77.8      & 79.5      & 82.9      & 86.3      & \B{67.6}  & 51.9      & 63.0      & 43.7      & 56.3      & 64.2\\
\hline
Our KS          & 15\%      & \B{58.7}  & \B{50.3}  & \B{62.0}  & \B{79.5}  & 75.4      & \B{84.9}  & 87.1      & 65.9      & 45.5      & 62.9      & 48.3      & 53.3      & \B{64.5}\\
Our KS          & 100\%     & 55.1      & 48.9      & 56.2      & 77.3      & \B{81.8}  & 83.3      & 86.4      & 62.6      & 51.9      & 61.3      & \B{51.0}  & \B{58.3}  & \B{64.5}\\
\hline
\end{tabular}
\begin{tabular}{l|c|*{8}{C{1.0cm}}|c}
\hline
Model          & Used labels & AU1       & AU2       & AU4       & AU6       & AU9       & AU12      & AU25      & AU26      & Avg.\\
\hline
DSIN            & 100\%     & 42.4      & 39.0      & 68.4      & 28.6      & 46.8      & 70.8      & 90.4      & 42.2      & 53.6\\
JAA            & 100\%     & 43.7      & 46.2      & 56.0      & 41.4      & 44.7      & 69.6      & 88.3      & 58.4      & 56.0\\
LP              & 100\%     & 29.9      & 24.7      & 72.7      & 46.8      & 49.6      & 72.9      & 93.8      & 65.0      & 56.9\\
ARL             & 100\%     & 43.9      & 42.1      & 63.6      & 41.8      & 40.0      & \B{76.2}  & \B{95.2}  & 66.8      & 58.7\\
SRERL           & 100\%     & 45.7      & 47.8      & 59.6      & 47.1      & 45.6      & 73.5      & 84.3      & 43.6      & 55.9\\ 
UGN             & 100\%     & 43.3      & 48.1      & 63.4      & 49.5      & 48.2      & 72.9      & 90.8      & 59.0      & 60.0\\
% SEV             & 100\%     & \B{55.3}  & 53.1      & 61.5      & 53.6      & 38.2      & 71.6      & \B{95.7}  & 41.5      & 58.8\\
HMP-PS          & 100\%     & 38.0      & 45.9      & 65.2      & 50.9      & \B{50.8}  & 76.0      & 93.3      & \B{67.6}  & 61.0\\
FAUDT           & 100\%     & 46.1      & 48.6      & \B{72.8 } & \B{56.7}  & 50.0      & 72.1      & 90.8      & 55.4      & 61.5\\
\hline
Our KS          & 15\%      & 41.7      & 53.5      & 69.7      & 41.3      & 46.2      & 72.0      & 92.3      & 54.0      & 58.8\\
Our KS          & 100\%     & \B{53.8}  & \B{59.9}  & 69.2      & 54.2      & \B{50.8 } & 75.8      & 92.2      & 46.8      & \B{62.8}\\
\hline
\end{tabular}
\end{table*}

\subsection{Effect of the Video Clip Length $n$}
\begin{figure}
\begin{center}
\includegraphics[width=1.0\linewidth]{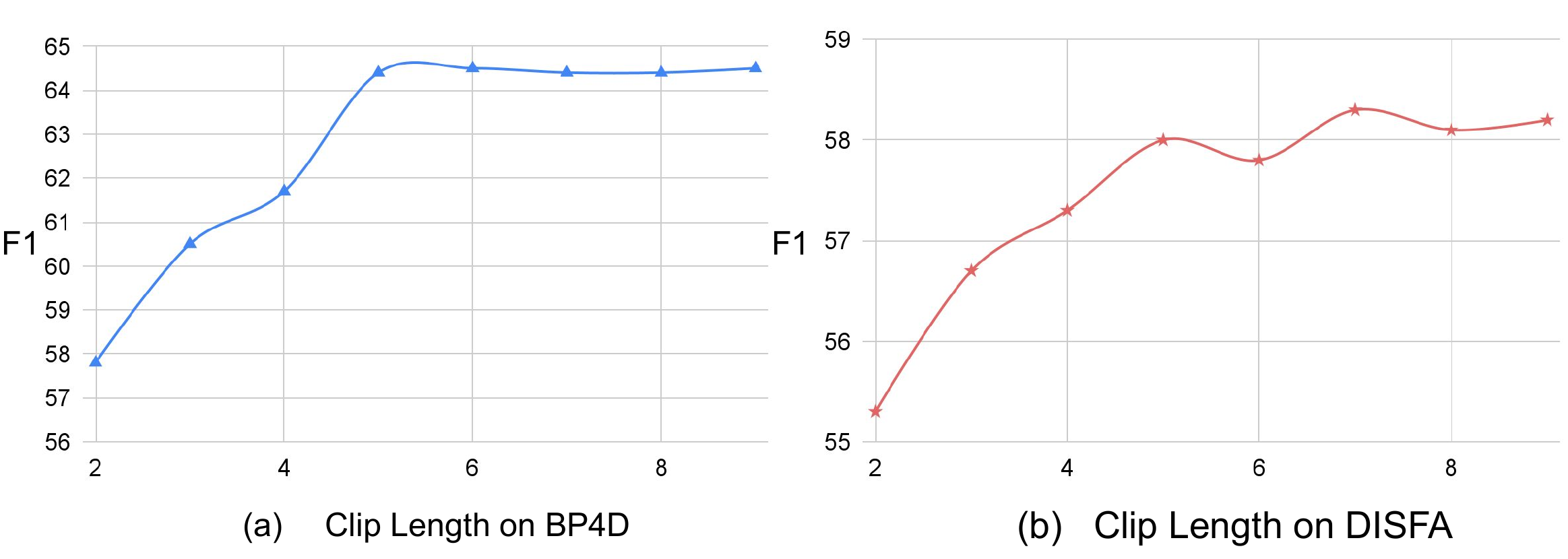}
\caption{Effect of the video clip length $n$. (a) and (b) indicates the F1 score with different video clip lengths on BP4D and DISFA.}
\label{hyper}
\end{center}
\end{figure}
\label{hyper-parameter}To investigate the influence of the input clip length, we perform experiments by the proposed model with 10\% sparsely sampled annotations on BP4D and DISFA. \Cref{hyper} shows the F1 score curve with $n$ changes. Overall, the performance improve with the $n$ increases from 2 to a certain threshold. A long video clip, on the other hand, results in high computational and memory costs. For optimal trading-off, 5 to 7 is a proper setting for the video clip length $n$.

\subsection{Parameter Scale Analysis}
The trainable parameter size of the proposed model is around 25 million, which makes KS a very light-weighted model. Compared with the baseline algorithm EACnet~\cite{Experiment_7}, which contains 138 million parameters, Knowledge-Spreader, as a video-level model, reduces considerable parameter (80\%) but achieves excellent performance improvement.

\section{Supplementary of New MME}
\subsection{Participants}
233 participants were recruited from our University. There are 132 females and 101 males, with ages ranging from 18 to 70 years old. Ethnic/Racial Ancestries include Asian, Black, Hispanic/Latino, White, and others (e.g., Native American). 

\subsection{Recording System and Synchronization}
Our data collection system consists of a 3D dynamic imaging camera system, a thermal sensor, a physiological signal sensor system, and a studio-quality audio recorder. The system setup and synchronization method are basically consistent with BP4D+~\cite{Experiments_17}.

\subsection{Emotion Stimulus}
\begin{table}\renewcommand\arraystretch{1}
\caption{ The stimulus tasks designed for the data collection.}
\begin{center}
\footnotesize
\resizebox{0.97\textwidth}{!}{
\begin{tabular}{c|c|c} \toprule
    {Task ID} & Activity & Target Emotion  \\ \midrule \midrule
     {1}           & Have a pleasant chat with the interviewer   &  Happiness    \\ \hline
     {2}                      & Watch a 3D face model of the participant   & Surprise    \\ \hline
    {3}                       & Watch an audio recording of 911 emergency call   & Sadness    \\ \hline
    {4}                       & Experience a sudden sound from a horn & Startle or Surprise \\ \hline
     {5} & React to a fake news & Skeptical  \\ \hline
      {6}               & Asked to sing an impromptu song & Embarrassment    \\ \hline
    {7}          & Experience physical fear of
the threat in a dart game  & Fear or Nervous \\ \hline
    {8}         & Experience the cold feeling by submerging \\ & hands into a bucket with ice water& Pain \\ \hline
    {9}              &  React to the blame from the interviewer & Offended or Unpleasant \\ \hline
    {10}              & Experience a bad smell from decaying food  & Disgust \\ 
    \bottomrule
\end{tabular}
}
\label{tab:tentask}
\end{center}
\end{table}
Ten tasks were performed to elicit a wide range of spontaneous emotion expression (from positive, to neutral, and to negative) and inter-personal facial action behavior by a professional interviewer. \Cref{tab:tentask} illustrates the detailed description for the designed tasks.

\subsection{Data Organization}
\begin{figure}
\begin{center}
\includegraphics[width=1.0\linewidth]{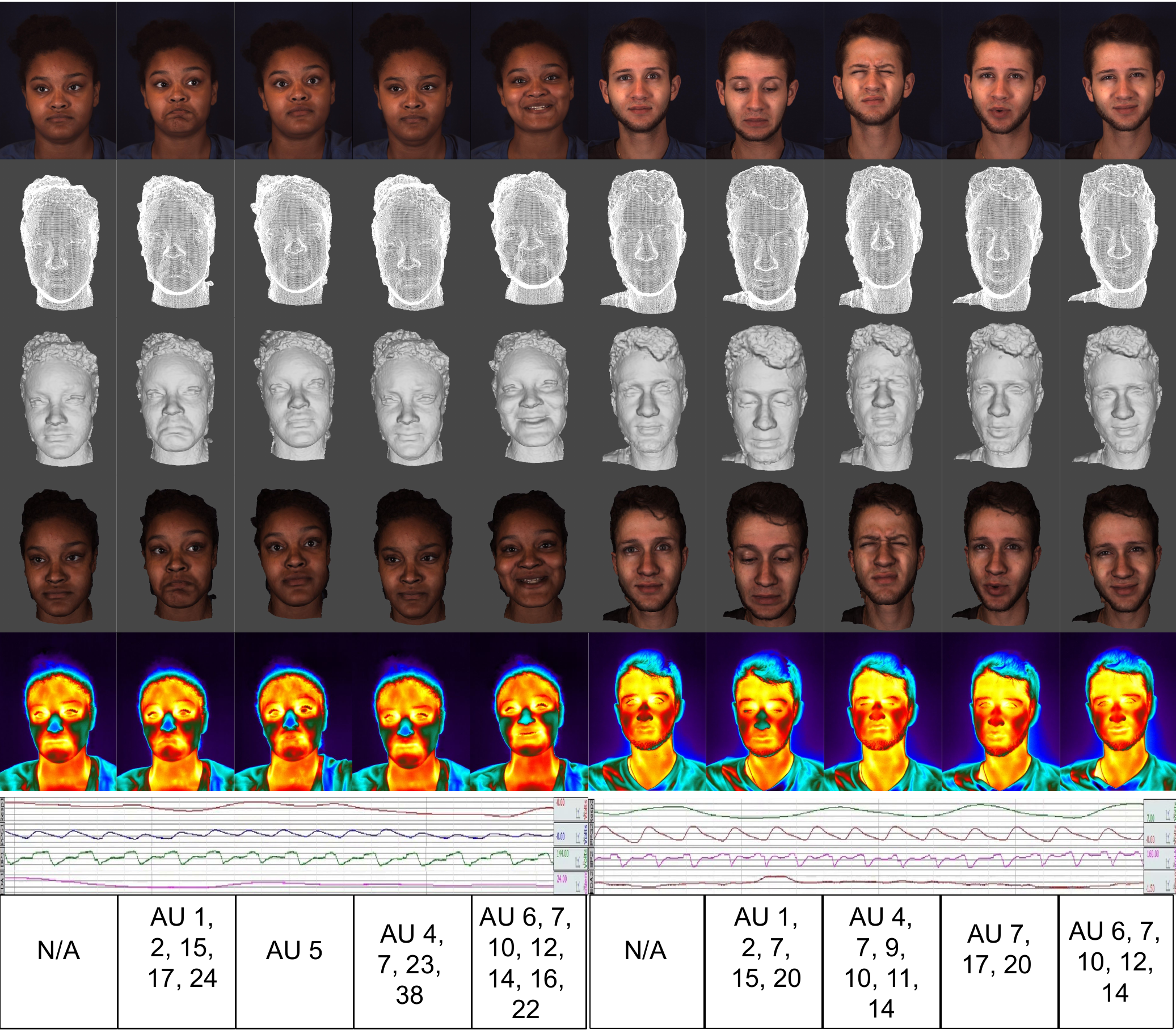}
\caption{A sample sequence from our MME. 2D texture image, 3D mesh model, 3D shaded model, 3D texture model, thermal image, and physiological signal (respiration rate, blood pressure, EDA, heart rate) and corresponding AU occurrence are shown from top to bottom.}
\label{dataset}
\end{center}
\end{figure}

Each subject is associated with 10 different emotions and multi-modal data including the 3D sequence, 2D RGB sequence, thermal sequence, and the sequences of physiological data (i.e., blood pressure, EDA, heart rate, and respiration rate). The sample sequences of different modalities from two subjects are shown in \Cref{dataset}. Besides, the metadata including manually labeled action units occurrence and intensity, 3D/2D/IR facial landmarks, and 3D head poses are also generated for better analysis of automatic human facial action.

\clearpage
\bibliographystyle{splncs04}
\bibliography{egbib}

\end{document}